\documentclass[review,3p,authoryear]{elsarticle}

\usepackage[utf8]{inputenc}
\usepackage[T1]{fontenc}
\usepackage{amsmath,amssymb,amsfonts}
\usepackage{graphicx}
\graphicspath{{figures/}}
\usepackage{booktabs}
\usepackage{array}
\usepackage{caption}
\usepackage{subcaption}
\usepackage{microtype}
\usepackage[colorlinks=true,linkcolor=blue,citecolor=blue,urlcolor=blue]{hyperref}
\usepackage{xcolor}
\usepackage{enumitem}
\usepackage{tikz}
\usetikzlibrary{arrows.meta,calc,positioning,fit,shapes.geometric}
\usepackage{lineno}

\journal{Computers \& Geosciences}

\begin{document}

\begin{frontmatter}

\title{DualTCN: A Physics-Constrained Temporal Convolutional Network
for Time-Domain Marine CSEM Inversion}

\author[siuc-cs]{Khaled~Ahmed\corref{cor1}}
\ead{khaled.ahmed@siu.edu}

\author[siuc-math]{Ghada~Omar}
\ead{ghada.omar@siu.edu}

\cortext[cor1]{Corresponding author.}

\affiliation[siuc-cs]{organization={School of Computing,
  Southern Illinois University Carbondale},
  city={Carbondale}, state={IL}, postcode={62901},
  country={USA}}

\affiliation[siuc-math]{organization={School of Mathematics and Statistics,
  Southern Illinois University Carbondale},
  city={Carbondale}, state={IL}, postcode={62901},
  country={USA}}

\begin{abstract}
We introduce DualTCN, to our knowledge the first deep-learning
framework for inverting time-domain marine controlled-source
electromagnetic (MCSEM) transient data. Instead of discretising the
subsurface, DualTCN regresses four earth-model
parameters---$\sigma_1$, $\sigma_2$, $d_1$, $d_2$---and reconstructs
the conductivity--depth profile through a differentiable soft-step
decoder. A systematic comparison of thirteen architectural variants
trained on one million \texttt{empymod} synthetics identifies DualTCN
(379\,K parameters) as the best design: a full-time temporal convolutional network (TCN) encoder
paired with a late-time branch (last 64 of 128 samples) and an
auxiliary seafloor-depth head yields a 25.3\% loss reduction over the
baseline, with $R^2 = 0.898$ ($\sigma_2$) and $0.627$ ($d_2$). A
single sample is inverted in 3.5\,ms on an A100 GPU. Training-time
noise augmentation confers robustness to waveform-channel noise at
SNR\,$\geq$\,10\,dB; however, the two-channel input design makes
$\sigma_2$ and $d_2$ critically dependent on the log-peak-amplitude
channel, so that $R^2_{d_2}$ turns negative at $\pm$2\% random
amplitude error per receiver, but a curriculum-based amplitude
augmentation strategy recovers full robustness ($\bar{R}^2 = 0.858$
at $\pm$2\% noise, versus $0.363$ without augmentation) while
preserving 98\% of clean-data accuracy.  A three-layer extension
(seawater/resistive layer/basement) generalises with no architectural
changes, resolving the basement conductivity at $R^2 \approx 0.88$,
though thin-layer thickness remains resolution-limited
($R^2 \approx 0.23$).
A multi-start benchmark shows DualTCN achieves
$\bar{R}^2 = 0.877$ versus $0.129$--$0.439$ for Levenberg--Marquardt
and L-BFGS-B (eight restarts) at up to $21{,}000\times$ lower cost.
Monte Carlo (MC) Dropout uncertainty quantification is well calibrated
for $\sigma_1$ (Prediction Interval Coverage Probability,
PICP$_{90} = 0.944$); the systematic under-coverage for
$d_2$ (PICP$_{90} = 0.572$) is correctable by post-hoc temperature
scaling or split conformal prediction and reflects limited signal
information at 200\,m offsets.
\end{abstract}


\begin{highlights}
\item First deep-learning framework for time-domain marine CSEM inversion
\item Dual-branch TCN with physics-constrained soft-step decoder
\item Ablation of 13 architectures on one million synthetic samples
\item Curriculum amplitude augmentation recovers robustness to 5\% noise
\item Inference in 3.5\,ms per sample on GPU with calibrated uncertainty
\end{highlights}

\end{frontmatter}

\section{Introduction}
\label{sec:intro}

Marine controlled-source electromagnetic (MCSEM) surveying has become
one of the standard geophysical tools for imaging the electrical
resistivity structure beneath the ocean floor. A horizontal electric
dipole (HED) source is towed near the seafloor, and an array of
receivers at varying offsets records the resulting electromagnetic
transient \citep{constable2006, constable2010ten}. The data carry
information about seawater conductivity, water depth, and the
resistivity of sub-seafloor formations---properties relevant to
hydrocarbon exploration, gas-hydrate characterisation, and studies of
oceanic crustal structure \citep{key2012marine}.

Turning these measurements into a subsurface model requires solving
an inverse problem. Conventional approaches start from an initial
guess and iteratively refine it by comparing forward-modelled
responses with the observed data. A single evaluation of an optimised
1D semi-analytic forward code such as \texttt{empymod}
\citep{werthmullerEmpymod} takes roughly 30--50\,ms on a modern
central processing unit (CPU). However, achieving a reliable solution requires hundreds to
thousands of such evaluations per sample: in our benchmark, a
multi-start Levenberg--Marquardt inversion (8 starts, up to
2\,000 evaluations each) consumes 12\,s per sample on average, and
L-BFGS-B with the same multi-start protocol takes 74\,s per sample
(Section~\ref{sec:benchmark}). Across a survey line of thousands of
records, the cumulative cost makes real-time quality control during
acquisition effectively impossible and constrains the throughput of
post-survey interpretation.

Deep learning offers an alternative that fundamentally changes this
cost structure \citep{lecun2015deep, reichstein2019deep}. Once a
neural network has been trained, inverting a new data sample requires
only a single feedforward pass, which takes milliseconds. This
concept of amortised inference has already been demonstrated for
several electromagnetic inverse problems
(Section~\ref{sec:related_work}), but important gaps remain. No
deep-learning method has been applied to time-domain MCSEM data. No
existing approach produces a physics-constrained parametric output.
And no architecture has been designed to exploit the fact that
different parts of the recorded transient carry information about
different subsurface parameters.

This paper introduces DualTCN, a deep-learning framework that
addresses each of these gaps. DualTCN operates directly on
multi-receiver time-domain transients, regresses four physically
meaningful earth-model scalars rather than a discretised depth
profile, and enforces the conductivity--depth structure through a
differentiable analytic decoder. We systematically compare thirteen
architectural variants trained on one million synthetic samples, and
show that a dual-branch temporal convolutional network with a
dedicated late-time encoder and an auxiliary seafloor-depth prediction
head achieves the best overall accuracy---with particular gains on
the source-to-seafloor distance $d_2$, the parameter that has proved most
resistant to improvement in every other configuration we tested.

\section{Background and Related Work}
\label{sec:related_work}

MCSEM data exhibit a temporal sensitivity hierarchy: early-time
samples sense mainly the water column ($\sigma_1$, $d_1$), while
late-time samples---where the signal decays as
$\exp(-2 u_1 d_2)$---carry information about the seafloor
($\sigma_2$, $d_2$) \citep{constable2010ten, key2012marine}.  This
hierarchy motivates DualTCN's dual-branch design
(Section~\ref{sec:dualtcn}).

Classical 1D inversion relies on Occam-style regularisation
\citep{constable1987occam, key2009} or gradient-based solvers, both
requiring dozens to hundreds of forward evaluations per sample---too
slow for real-time use.  Deep learning offers amortised inference:
CNN-based frequency-domain CSEM inversion
\citep{puzyrev2019deep, puzyrev2021multi, zhang2024_3d}, neural-network
initial models that accelerate conventional solvers
\citep{araya2018deep}, and physics-guided ATEM strategies
(disentangled encoders, auxiliary depth losses, differentiable
decoders) \citep{moghadas2020deep, liu2021atem, colombo2021deep}.
Alternative paradigms include projected Gauss--Newton with learned
priors \citep{abubakar2012pgn}, Deep Image Prior for MT
\citep{sun2023dip_mt, bai2024diffmt}, ensemble Kalman inversion
\citep{iglesias2013eki}, and probabilistic methods (invertible neural
networks, normalising flows, deep ensembles)
\citep{ardizzone2019inn, papamakarios2021normalizing, lakshminarayanan2017}.

Despite this progress, six gaps remain (see Supplement Section~S11
for detailed discussion): (1)~no DL method has addressed time-domain
marine CSEM; (2)~all existing methods predict discretised profiles
rather than physics-constrained parametric outputs; (3)~no systematic
encoder comparison (TCN vs.\ Transformer vs.\ MLP) exists for this
setting; (4)~no architecture dedicates a branch to the late-time
regime or uses auxiliary physical objectives for $d_2$;
(5)~ATEM physics-guided strategies have not been transferred to
marine CSEM; (6)~no calibrated UQ has been demonstrated.
DualTCN addresses gaps 1--4 directly, draws on ATEM methodology for
gap~5, and provides MC-Dropout UQ with post-hoc calibration for
gap~6.

\section{Methods}
\label{sec:method}

\subsection{Earth Model and Synthetic Data}
\label{sec:params}

\begin{figure}[t]
  \centering
  \begin{tikzpicture}[scale=0.85, >=Stealth, font=\scriptsize]
    \fill[gray!12]   (0,0)    rectangle (6.2, 1.0);
    \fill[cyan!18]   (0,-4.2) rectangle (6.2, 0);
    \fill[brown!30]  (0,-5.5) rectangle (6.2,-4.2);
    \draw[blue!55, thick]        (0,0)    -- (6.2, 0);
    \draw[brown!70!black, thick] (0,-4.2) -- (6.2,-4.2);
    \node[gray!55!black, font=\scriptsize] at (1.0, 0.6) {Air};
    \node[gray!55!black, font=\scriptsize] at (1.0, 0.2)
      {$\rho \to \infty$};
    \node[cyan!40!black, align=center] at (1.0,-1.5)
      {Sea-\\water\\$\sigma_1$};
    \node[brown!65!black, align=center] at (1.0,-4.9)
      {Seafloor\\$\sigma_2$};
    \node[blue!55, right, font=\scriptsize] at (6.2, 0.08)
      {$z = 0$};
    \draw[red!75!black, very thick, ->] (2.1,-3.0) -- (2.55,-3.0);
    \draw[red!75!black, very thick, ->] (3.3,-3.0) -- (2.85,-3.0);
    \fill[red!75!black] (2.7,-3.0) circle (3pt);
    \node[red!75!black, above, font=\scriptsize] at (2.7,-2.87)
      {HED};
    \draw[<->, red!60!black] (-0.35, 0) -- (-0.35,-3.0);
    \node[left, red!60!black, font=\scriptsize] at (-0.38,-1.5)
      {$d_1$};
    \draw[<->, red!60!black] (-0.35,-3.0) -- (-0.35,-4.2);
    \node[left, red!60!black, font=\scriptsize] at (-0.38,-3.6)
      {$d_2$};
    \foreach \rx/\rlbl in {3.0/20, 3.9/50, 5.0/100, 6.0/200}{
      \fill[green!50!black]
        (\rx-0.13,-0.18) -- (\rx+0.13,-0.18) --
        (\rx,-0.50) -- cycle;
      \node[green!50!black, above, font=\tiny] at (\rx,-0.12)
        {\rlbl\,m};
    }
    \node[green!50!black, font=\scriptsize] at (4.6, 0.75)
      {Receivers ($z_\mathrm{obs} = 20$\,m)};
    \draw[orange!65!black, dashed, thin]
      (2.7,-3.0) -- (3.0,-0.50);
    \draw[orange!65!black, dashed, thin]
      (2.7,-3.0) -- (3.9,-0.50);
    \draw[purple!60!black, dashed, thin]
      (2.7,-3.0) -- (2.7,-4.2) -- (5.0,-4.2) -- (5.0,-0.50);
    \draw[purple!60!black, dashed, thin]
      (2.7,-3.0) -- (2.7,-4.2) -- (6.0,-4.2) -- (6.0,-0.50);
    \node[font=\bfseries\footnotesize] at (0.35, 0.85) {(a)};
    \def\px{7.1}
    \def\pw{1.7}
    \fill[gray!6] (\px,-5.5) rectangle (\px+\pw,1.0);
    \draw[gray!50] (\px,-5.5) rectangle (\px+\pw,1.0);
    \node[font=\scriptsize, rotate=90] at (\px-0.35,-2.25)
      {Depth $z$};
    \node[font=\scriptsize] at (\px+0.85,1.25) {$\sigma(z)$};
    \draw[blue!40, thin, dashed] (\px,0) -- (\px+\pw,0);
    \draw[brown!50, thin, dashed] (\px,-4.2) -- (\px+\pw,-4.2);
    \draw[blue!60!black, very thick]
      (\px+0.90, 0) -- (\px+0.90,-4.2)
      -- (\px+0.30,-4.2) -- (\px+0.30,-5.5);
    \node[blue!60!black, right, font=\scriptsize]
      at (\px+0.92,-2.0) {$\sigma_1$};
    \node[blue!60!black, right, font=\scriptsize]
      at (\px+0.32,-4.9) {$\sigma_2$};
    \node[font=\bfseries\footnotesize] at (\px+0.2, 0.85) {(b)};
    \draw[orange!65!black, dashed, thin]
      (0.2,-5.2) -- (0.9,-5.2);
    \node[right, font=\tiny] at (0.9,-5.2)
      {Direct (near offset)};
    \draw[purple!60!black, dashed, thin]
      (3.2,-5.2) -- (3.9,-5.2);
    \node[right, font=\tiny] at (3.9,-5.2)
      {Refracted (far offset)};
  \end{tikzpicture}
  \caption{(a)~MCSEM acquisition geometry. The HED source sits at
  depth $d_1$ below the sea surface ($z = 0$); the seafloor interface
  is a further $d_2$ below the source, giving a total seafloor depth
  $d_\mathrm{sf} = d_1 + d_2$. Four receivers span offsets of
  20--200\,m. Orange dashed lines show the direct path; purple dashed
  lines show the seafloor-refracted path. (b)~Step-function
  conductivity profile $\sigma(z)$ reconstructed by the physics
  decoder from four predicted scalars. The seafloor half-space
  ($\sigma_2$) extends to infinite depth.}
  \label{fig:earthmodel}
\end{figure}
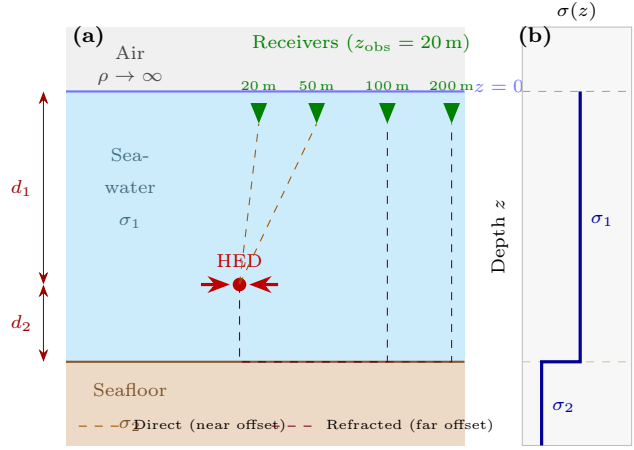

We use a one-dimensional three-layer earth model: air
($\rho_\mathrm{air} \to \infty$), seawater (conductivity $\sigma_1$),
and a seafloor half-space (conductivity $\sigma_2$, extending
to infinite depth). The geometry is shown in
Fig.~\ref{fig:earthmodel}. A unit-moment HED source is placed at
depth $d_1$ below the sea surface (i.e., at a height $d_2$ above
the seafloor interface, where $d_2 = d_\mathrm{sf} - d_1$). The
total seafloor depth is therefore $d_\mathrm{sf} = d_1 + d_2$.
To be explicit: $d_1$ is the vertical distance from the sea surface
to the HED source, and $d_2$ is the vertical distance from the source
down to the seafloor interface. The seafloor is modelled as a uniform
half-space of conductivity $\sigma_2$ extending to infinite depth;
$d_2$ parametrises the gap between source and seafloor, not the
thickness of a finite resistive layer. Four inline receivers at
offsets of 20, 50, 100, and 200\,m record the transient electric
field. Receivers are modelled at a fixed observation depth of
$z_\mathrm{obs} = 20$\,m below the sea surface (i.e., in the water
column, not on the seafloor).

The four earth-model parameters are sampled independently and
uniformly within the ranges given in Table~\ref{tab:params}. All
training and prediction are carried out in $\log_{10}$ space. One
million parameter combinations are drawn, and the corresponding
forward responses are computed with \texttt{empymod}
\citep{werthmullerEmpymod}, evaluating the response at 64 linearly
spaced frequencies from 0.05 to 2.0\,Hz and transforming to the time
domain via \texttt{numpy.fft.irfft} with output length $N = 128$
($\Delta t = 0.25$\,s, 0--32\,s window).  The 0.05--2.0\,Hz band
spans the skin-depth range relevant to our geometry
($\delta \approx 205$--$1{,}300$\,m in seawater).  The synthesis
uses the impulse-response convention; deployment on step-off data
requires a frequency-domain pre-processing step validated in
Supplement Section~S2.  Three known simplifications---absent DC
component, 2\,Hz bandwidth truncation, and fixed receiver
geometry---are discussed in detail in Supplement Section~S12; the
pipeline is self-consistent, ensuring accuracy metrics reflect
genuine earth-model sensitivity (Supplement Section~S1 confirms
$r \geq 0.9998$ versus a corrected-DC reference).

The one million samples are split 70/15/15 into training (700\,000),
validation (150\,000), and test (150\,000) subsets.
The training set is augmented with additive Gaussian
noise (Section~\ref{sec:noise}); the validation and test sets are
evaluated on clean data unless otherwise noted.

Each receiver trace is normalised by its peak amplitude; the
$\log_{10}$ of that peak is appended as a second channel, preserving
the absolute signal level.  The normalised waveform carries timing
and decay information (sensitive to $\sigma_1$, $d_1$), while the
log-amplitude channel is the primary discriminator for $\sigma_2$ and
$d_2$.  This creates a known vulnerability: any uncompensated
amplitude error directly corrupts $\sigma_2$ and $d_2$
(Section~\ref{sec:amp_noise}).
The resulting input tensor has dimensions
$(8, 128)$: four receivers $\times$ two channels by 128 time samples.

\begin{table}[t]
  \centering
  \footnotesize
  \caption{Earth-model parameter ranges for synthetic data
  generation. $d_1$ is the depth of the HED source below the sea
  surface; $d_2$ is the vertical distance from the source down to
  the seafloor interface ($d_\mathrm{sf} = d_1 + d_2$). The seafloor
  is a uniform half-space; $\sigma_2$ applies from $d_\mathrm{sf}$
  to infinite depth.}
  \label{tab:params}
  \begin{tabular}{lcc}
    \toprule
    Parameter & Physical range & $\log_{10}$ range \\
    \midrule
    $\sigma_1$ (S\,m$^{-1}$) & 0.10--5.01   & $-$1.00 to 0.70 \\
    $\sigma_2$ (S\,m$^{-1}$, half-space) & 0.001--1.0   & $-$3.00 to 0.00 \\
    $d_1$ (m, source depth) & 50--150      & 1.70--2.18 \\
    $d_2$ (m, source-to-seafloor) & 10--50       & 1.00--1.70 \\
    \bottomrule
  \end{tabular}
\end{table}

\begin{table}[t]
  \centering
  \footnotesize
  \caption{Notation and symbols used throughout this paper.}
  \label{tab:notation}
  \begin{tabular}{lp{7cm}}
    \toprule
    Symbol & Definition \\
    \midrule
    $\sigma_1$ & Seawater conductivity (S\,m$^{-1}$) \\
    $\sigma_2$ & Seafloor half-space conductivity (S\,m$^{-1}$) \\
    $d_1$      & Depth of HED source below the sea surface (m) \\
    $d_2$      & Vertical distance from source to seafloor interface (m) \\
    $d_\mathrm{sf}$ & Total seafloor depth: $d_\mathrm{sf} = d_1 + d_2$ (m) \\
    $\hat{d}_\mathrm{sf}$ & Auxiliary predicted seafloor depth (DualTCN training target) \\
    $z_\mathrm{obs}$ & Receiver observation depth below sea surface (20\,m) \\
    HED        & Horizontal Electric Dipole (source) \\
    $N_T$      & Number of time samples per trace (128) \\
    $N_R$      & Number of inline receivers (4, at offsets 20--200\,m) \\
    $\bar{R}^2$ & Mean $R^2$ across all four parameters \\
    PICP$_\alpha$ & Prediction Interval Coverage Probability at nominal level $\alpha$ \\
    MPIW       & Mean Prediction Interval Width \\
    \bottomrule
  \end{tabular}
\end{table}

\subsection{Baseline Architecture}
\label{sec:baseline}

The baseline model, which we call PCRN (Physics-Constrained Parameter
Regression Network), has 638\,000 trainable parameters and three
stages: a feature encoder, a parameter prediction head, and a
differentiable physics decoder.

\subsubsection{Hybrid Encoder}

The encoder has both a local and a global component. The local
component consists of six dilated residual blocks with dilation
factors that double at each stage (1, 2, 4, 8, 16, 32). Each block
applies a dilated causal convolution, batch normalisation, Gaussian
Error Linear Unit (GELU) activation, and a residual connection. The exponential dilation
schedule gives the network a receptive field spanning the full
128-sample input while keeping the parameter count modest. The global
component is a two-layer Transformer encoder with four attention
heads and pre-layer normalisation, which captures dependencies across
the entire temporal extent. Adaptive average pooling compresses the
Transformer output into a latent vector
$\mathbf{z} \in \mathbb{R}^{256}$.

\subsubsection{Parameter Head}

A two-layer MLP ($256 \to 128 \to 4$) with sigmoid output maps the
latent vector to normalised predictions
$\hat{\mathbf{p}} \in [0,1]^4$, which are then rescaled to the
physical $\log_{10}$ ranges in Table~\ref{tab:params}.

\subsubsection{Physics Decoder}

The decoder converts the four predicted scalars into a
conductivity--depth profile using a differentiable analytic function.
Because the parameter head outputs normalised $[0,1]$ values that are
linearly interpolated to $\log_{10}$ ranges (Section~\ref{sec:params},
Table~\ref{tab:params}), the decoder first exponentiates each
predicted quantity back to physical units before evaluation:
\begin{align}
  \tilde{\sigma}_k &= 10^{\hat{p}_{\sigma_k}}
      \quad [{\rm S\,m^{-1}}], \quad k \in \{1, 2\}, \\
  \tilde{d}_k      &= 10^{\hat{p}_{d_k}}
      \quad [{\rm m}],           \quad k \in \{1, 2\}.
  \label{eq:decode_units}
\end{align}
The seafloor depth is then the arithmetic sum of the two linear-space
depths, $\hat{d}_\mathrm{sf} = \tilde{d}_1 + \tilde{d}_2$\,(m), and
the conductivity profile is (see also Table~\ref{tab:params})
\begin{equation}
  \sigma(z) = \tilde{\sigma}_1
    + (\tilde{\sigma}_2 - \tilde{\sigma}_1)\,
    \sigma_\mathrm{s}\!\left(
      \frac{z - \hat{d}_\mathrm{sf}}{\tau}
    \right),
  \quad \tau = 2\,\mathrm{m},
  \label{eq:profile}
\end{equation}
where $\sigma_\mathrm{s}$ is the logistic sigmoid, $z$ is depth in
metres, and $\sigma(z)$ is in S\,m$^{-1}$. The decoder evaluates
$\log_{10}\sigma(z)$ at $N_z = 64$ uniformly spaced depth nodes
spanning $z \in [0, 250]$\,m ($\Delta z = 250/63 \approx 3.97$\,m),
which matches the maximum parameter depth ($d_1^{\max}+d_2^{\max} =
150+50 = 200$\,m) with a $50$\,m buffer. Note that the transition
half-width $\tau = 2$\,m $\approx \Delta z/2$, so the sigmoid
transition is resolved by at least two depth samples. Because it is
fully differentiable, gradients from the profile loss propagate back
through the exponentiation and the predicted parameters into the
encoder, enforcing the two-layer shape as a structural regulariser.

\paragraph{Sensitivity to $\tau$.}
Because the four parameter predictions ($\hat{\sigma}_1$,
$\hat{\sigma}_2$, $\hat{d}_1$, $\hat{d}_2$) are produced by the
regression heads \emph{before} the decoder, they are architecturally
independent of $\tau$: the decoder only consumes these predictions to
construct the profile.  We verified this by evaluating the trained
model at inference time with $\tau \in \{0.5, 1, 2, 4, 8\}$\,m on
150\,000 test samples.  All parameter $R^2$ values are identical
across all $\tau$ values (e.g., $R^2_{\sigma_2} = 0.898$ and
$R^2_{d_2} = 0.627$ for every $\tau$), and the mean bias in
$\log_{10}\hat{\sigma}_2$ and $\log_{10}\hat{d}_2$ is likewise
invariant ($+0.073$ and $-0.025$, respectively).  Only the profile
MSE changes: it is minimised near $\tau = 2$\,m against the
hard-step ground truth (MSE\,$= 0.065$) and decreases monotonically
with larger $\tau$ when evaluated against a matched-$\tau$ reference.
Thus, $\tau$ is a visualization hyperparameter that controls the
smoothness of the reconstructed profile but introduces \textbf{no
bias} in the inversion parameters.

\subsubsection{Loss Function and Training}

The loss combines profile reconstruction with per-parameter
regression:
\begin{equation}
  \mathcal{L} = \mathcal{L}_\mathrm{MSE}^\mathrm{prof}
    + \sum_{i=1}^{4} w_i \,
      \mathcal{L}_\mathrm{Huber}^{(i)},
  \label{eq:loss}
\end{equation}
with Huber $\delta = 0.1$ and weights
$\mathbf{w} = [1, 3, 3, 2]$.
The profile MSE $\mathcal{L}_\mathrm{MSE}^\mathrm{prof}$ is the
unweighted mean over all 64 depth nodes; no depth-dependent weighting
is applied, so shallow and deep nodes contribute equally per unit
depth interval (uniform spacing ensures equal representation). Training uses AdamW
($\lambda = 10^{-3}$, weight decay $10^{-3}$), a one-cycle cosine
schedule (5-epoch linear warm-up, peak rate $5 \times 10^{-4}$,
final rate $5 \times 10^{-6}$), batch size 256, for 100 epochs on
the 700\,000-sample training set (approximately 2\,734 gradient
steps per epoch) on a single NVIDIA A100 40\,GB GPU. Training
DualTCN for 100 epochs required approximately 3--4 GPU-hours on this
hardware; all training used PyTorch 2.x with automatic mixed
precision (AMP) and \texttt{torch.compile} for graph optimisation.
No explicit early stopping is applied; instead, the model
checkpoint with the lowest total validation loss across all 100
epochs is saved and used for all subsequent evaluation. All variants
share this protocol and are evaluated on the held-out
150\,000-sample test set.  Table~\ref{tab:compute} reports the
computational cost for the key variants.

\begin{table}[t]
  \centering
  \footnotesize
  \caption{Computational cost for key DualTCN variants.
  GPU = NVIDIA A100 40\,GB; CPU = single core (OMP\_NUM\_THREADS=1).}
  \label{tab:compute}
  \setlength{\tabcolsep}{2pt}
  \begin{tabular}{lrrrrr}
    \toprule
    Variant & Params & Train & \multicolumn{2}{c}{Inference (ms)} & Thru. \\
    \cmidrule(lr){4-5}
            &        & (GPU-h) & GPU & CPU & (K/s) \\
    \midrule
    PCRN (baseline)     & 638\,K & 4.0 & 2.6 & 12.1 &  45 \\
    P5a (TCN-only)      & 201\,K & 2.5 & 2.0 &  6.5 & 127 \\
    P8 (TwoStage)       & 401\,K & 3.5 & 3.8 & 10.8 &  67 \\
    P10a (iTransf.)     & 381\,K & 3.0 & 1.0 &  5.2 & 247 \\
    \textbf{DualTCN}    & 379\,K & 3.5 & 3.5 &  8.8 &  76 \\
    DualTCN-AmpAug      & 379\,K & 3.5 & 3.5 &  8.8 &  76 \\
    DualTCN-3Layer      & 306\,K & 3.5 & 3.5 &  8.5 &  76 \\
    \bottomrule
  \end{tabular}
\end{table}

\section{Architectural Ablation}
\label{sec:ablation}

We ran thirteen parallel experiments, each modifying one aspect of
the PCRN baseline (Supplementary Section~S7 lists all variants).
The key intermediate designs are P7 (multi-scale TCN: three parallel
branches with different dilation schedules targeting early-, mid-,
and late-time features; 384\,K parameters) and P8 (two-stage
cascaded TCN: Stage~1 predicts $[\sigma_1, d_1]$, Stage~2
conditions on gradient-detached Stage~1 outputs to predict
$[\sigma_2, d_2]$; 401\,K parameters).  Two attention-based
variants---P10a (iTransformer, treating each channel as a token) and
P10b (PatchTST, segmenting traces into overlapping patches)---were
also evaluated; both underperform all TCN designs
(Section~\ref{sec:transformer_discussion}).

\subsection{DualTCN: Late-Time Branch with Auxiliary Depth Head}
\label{sec:dualtcn}

\begin{figure*}[!t]
  \centering
  \includegraphics[width=\textwidth,height=0.35\textheight,keepaspectratio=false]{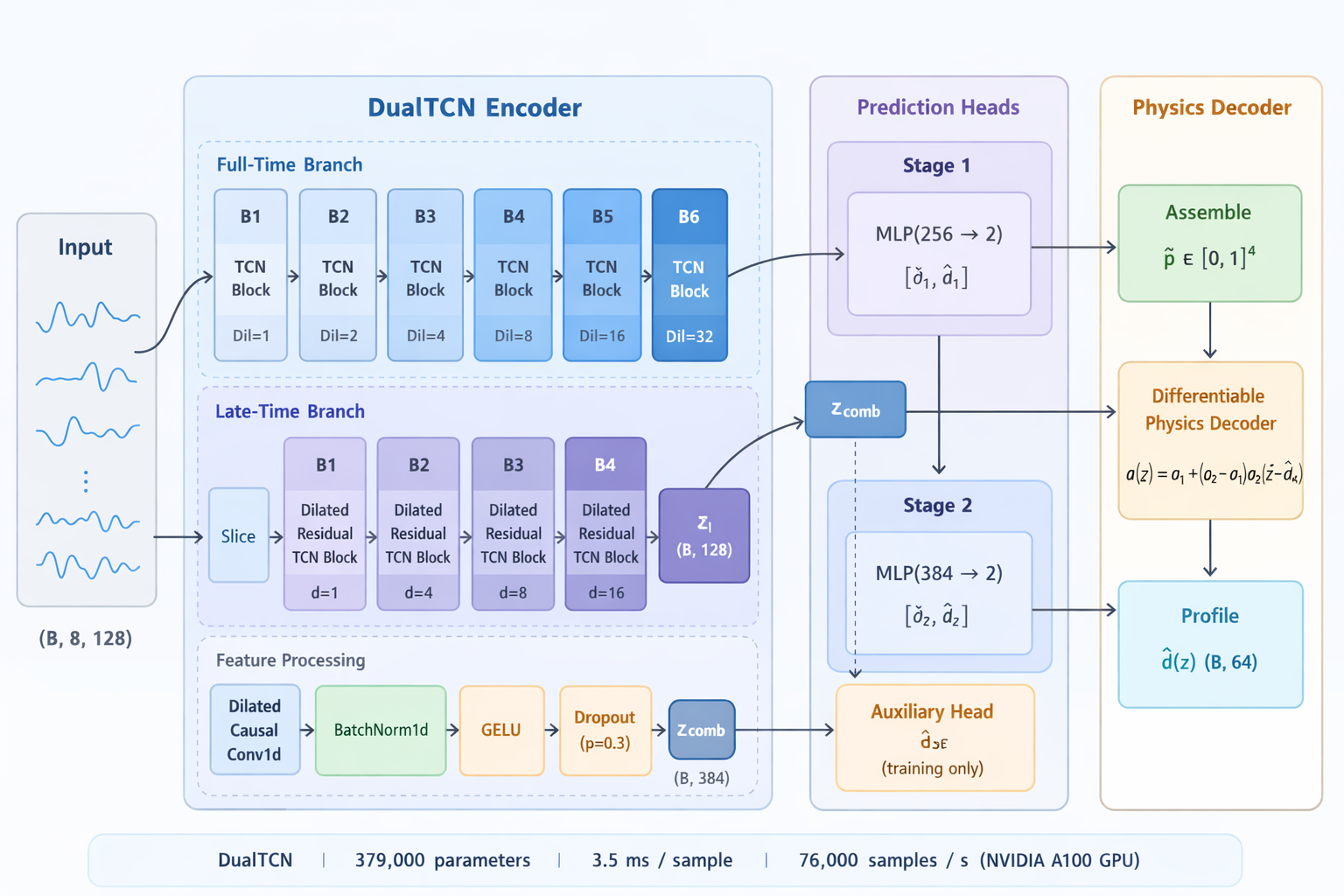}
  \caption{Architecture of DualTCN (379\,000 parameters).
  \emph{Full-Time Branch} (top): six TCN blocks (B1--B6) with
  exponentially growing dilations 1, 2, 4, 8, 16, 32 process the
  full 128-sample input and produce the latent
  $\mathbf{z}_\mathrm{f} \in \mathbb{R}^{256}$.
  \emph{Late-Time Branch} (middle): the input is first sliced to the
  last 64 samples; four Dilated Residual TCN blocks (dilations 1, 4,
  8, 16) encode this late-time window into
  $\mathbf{z}_\mathrm{l} \in \mathbb{R}^{128}$.  The two latents are
  concatenated to form $\mathbf{z}_\mathrm{comb} \in \mathbb{R}^{384}$.
  \emph{Feature Processing zoom} (bottom inset): internal structure
  of every TCN block---a causal dilated convolution followed by
  Batch\-Norm\-1d, GELU activation, and Dropout ($p = 0.30$).
  \emph{Prediction Heads}: Stage~1 maps
  $\mathbf{z}_\mathrm{f}$ through an MLP ($256 \to 2$) to
  $[\hat{\sigma}_1, \hat{d}_1]$; Stage~2 maps
  $\mathbf{z}_\mathrm{comb}$ conditioned on gradient-detached
  Stage~1 outputs through an MLP ($384 \to 2$) to
  $[\hat{\sigma}_2, \hat{d}_2]$.  An auxiliary head predicts
  $\hat{d}_\mathrm{sf}$ during training only (yellow box).
  \emph{Physics Decoder}: the four predicted scalars are assembled
  into $\hat{\mathbf{p}} \in [0,1]^4$ and passed through the
  differentiable soft-step decoder to produce the conductivity
  profile $\hat{\sigma}(z) \in \mathbb{R}^{64}$.}
  \label{fig:p9arch}
\end{figure*}

DualTCN builds on P7 and P8 by adding two mechanisms aimed
specifically at improving $d_2$ (Fig.~\ref{fig:p9arch}).

The first is a dedicated late-time encoder: a compact four-block
dilated TCN (32 channels, dilations 1, 4, 8, 16) applied only to the
last 64 of 128 time samples---the window dominated by the
exponential decay $E(t) \sim \exp(-2 u_1 d_2)$. Its 128-dimensional
latent $\mathbf{z}_\mathrm{late}$ is concatenated with the full-time
encoder's 256-dimensional latent to form
$\mathbf{z}_\mathrm{comb} \in \mathbb{R}^{384}$.

The second is an auxiliary seafloor-depth head: a small MLP that
predicts $d_\mathrm{sf} = d_1 + d_2$ (range 60--200\,m). Because
the data are more sensitive to $d_\mathrm{sf}$ than to $d_2$ alone,
predicting it steers the combined latent toward features useful for
$d_2$. The auxiliary Huber loss (weight 0.5) supplements
Eq.~\ref{eq:loss} during training only and is discarded at test
time. Total parameters: 379\,000.

Head~1 maps $\mathbf{z}_\mathrm{full}$ to the easy parameters
$[\hat{\sigma}_1, \hat{d}_1]$. Head~2 receives
$[\mathbf{z}_\mathrm{comb};\, \hat{\sigma}_1^\perp;\,
\hat{d}_1^\perp]$ (gradient-detached Stage~1 outputs) and predicts
$[\hat{\sigma}_2, \hat{d}_2]$. The assembled $\hat{\mathbf{p}}$
feeds the physics decoder (Eq.~\ref{eq:profile}).

\section{Results}
\label{sec:results}

\subsection{Ablation Results}

Table~\ref{tab:results} reports normalised root-mean-square error (RMSE),
$R^2$, and inference timing for all variants. All RMSE values are in
the normalised $[0,1]$ log-space used during training. All $R^2$
values are likewise computed on parameters normalised to $[0,1]$ in
$\log_{10}$ space; they therefore measure explained variance in
$\log_{10}$-parameter space, not in physical units.
Supplementary Table~S4 (Section~\ref{sec:pred_quality}) provides
a systematic conversion of all four DualTCN RMSE values to physical
units (S\,m$^{-1}$ and m), including the geometric RMSE factor and
representative absolute errors at the geometric mean of each training
range.

\begin{table*}[t]
  \centering
  \caption{Ablation summary: normalised RMSE, $R^2$, and latency for
  the baseline, top-performing variants, and representative
  underperformers (A100 GPU). Full results for all 13 variants are in
  Supplementary Tables~S1--S2.  Bold = best.}
  \label{tab:results}
  \small
  \setlength{\tabcolsep}{3.5pt}
  \begin{tabular}{llrcccccccr}
    \toprule
    & Variant & Params & \multicolumn{4}{c}{RMSE (norm.)} & Loss
      & $R^2_{d_2}$ & $\bar{R}^2$ & ms \\
    \cmidrule(lr){4-7}
    & & & $\sigma_1$ & $\sigma_2$ & $d_1$ & $d_2$ & & & & \\
    \midrule
    & PCRN (baseline) & 638K & 0.026 & 0.110 & 0.033 & 0.183
      & 0.103 & 0.541 & 0.844 & 2.6 \\
    P5a & TCN-only    & \textbf{201K} & 0.022 & 0.109 & 0.031
      & 0.187 & 0.102 & 0.525 & 0.841 & 2.0 \\
    P7  & MultiScale  & 384K & 0.021 & 0.100 & 0.031 & 0.184
      & 0.093 & 0.539 & 0.850 & 3.8 \\
    P8  & TwoStage    & 401K & 0.020 & 0.098 & 0.033 & 0.185
      & 0.089 & 0.532 & 0.852 & 3.8 \\
    P10a & iTransf.   & 381K & 0.040 & 0.180 & 0.041 & 0.220
      & 0.220 & 0.342 & 0.728 & 1.0 \\
    \textbf{DualTCN}
      &              & 379K & \textbf{0.020} & \textbf{0.092}
      & \textbf{0.031} & \textbf{0.165}
      & \textbf{0.077} & \textbf{0.627} & \textbf{0.877} & 3.5 \\
    \bottomrule
  \end{tabular}
\end{table*}

Three patterns stand out. First, the choice of encoder architecture
matters more than any other design decision. The Transformer-only
(P5b), MLP-only (P5c), iTransformer (P10a), and PatchTST (P10b)
variants all perform far worse than any TCN-based model. Second,
among the TCN-based designs, multi-scale branching (P7), hierarchical
conditioning (P8), and late-time specialisation (DualTCN) produce a
monotonic improvement: total losses of 0.0931, 0.0892, and 0.0771.
Third, DualTCN achieves the lowest loss and the highest $R^2$ for
every parameter, including $R^2_{d_2} = 0.627$---up from 0.532 in
P8.

\subsection{Prediction Quality}
\label{sec:pred_quality}

\begin{figure}[t]
  \centering
  \includegraphics[width=\columnwidth]{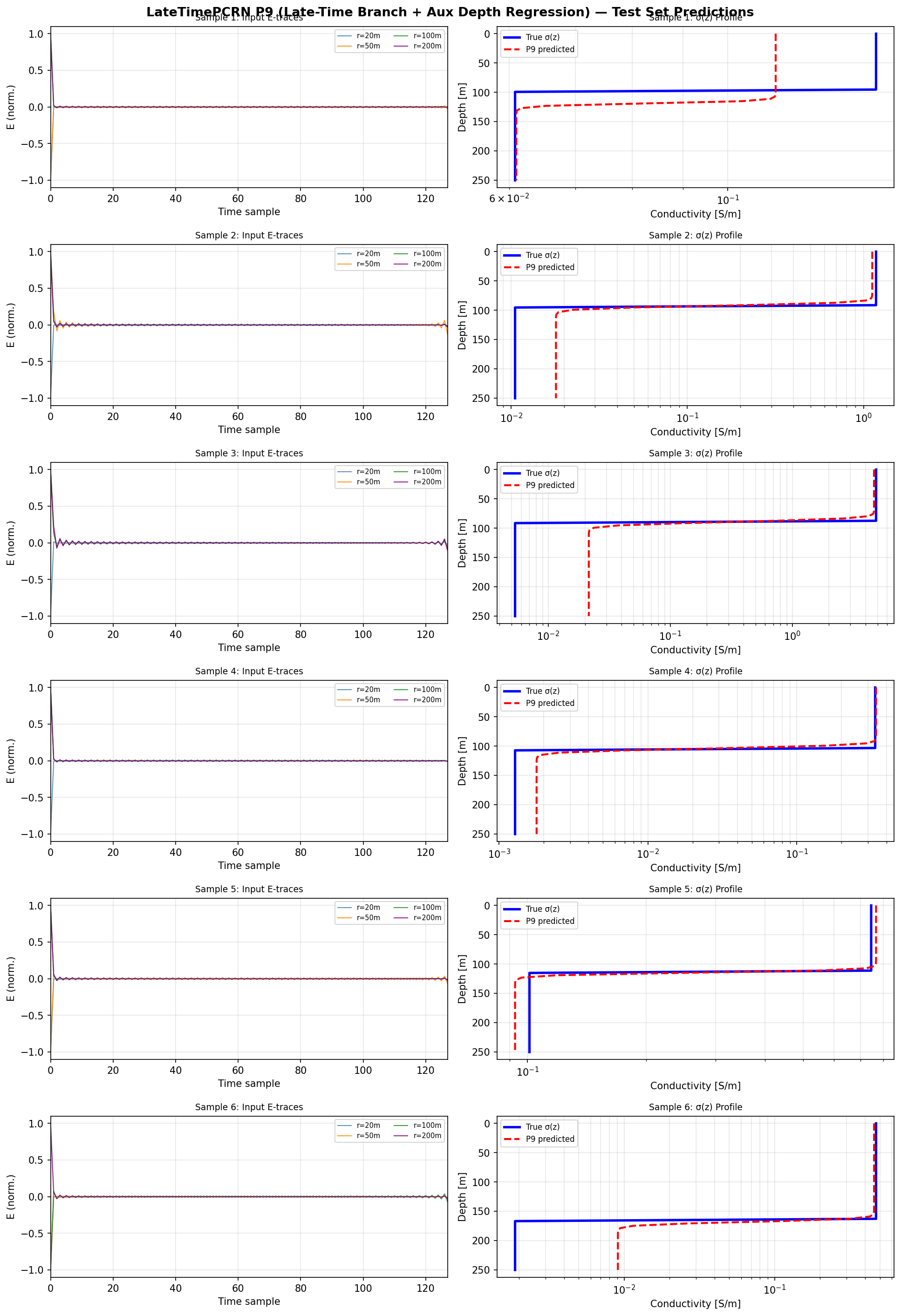}
  \caption{DualTCN predictions on six test samples.  Left: normalised
  E-field traces from four receivers.  Right: true profile (blue) and
  prediction (red dashed).}
  \label{fig:predictions}
\end{figure}

\begin{figure}[t]
  \centering
  \includegraphics[width=\columnwidth]{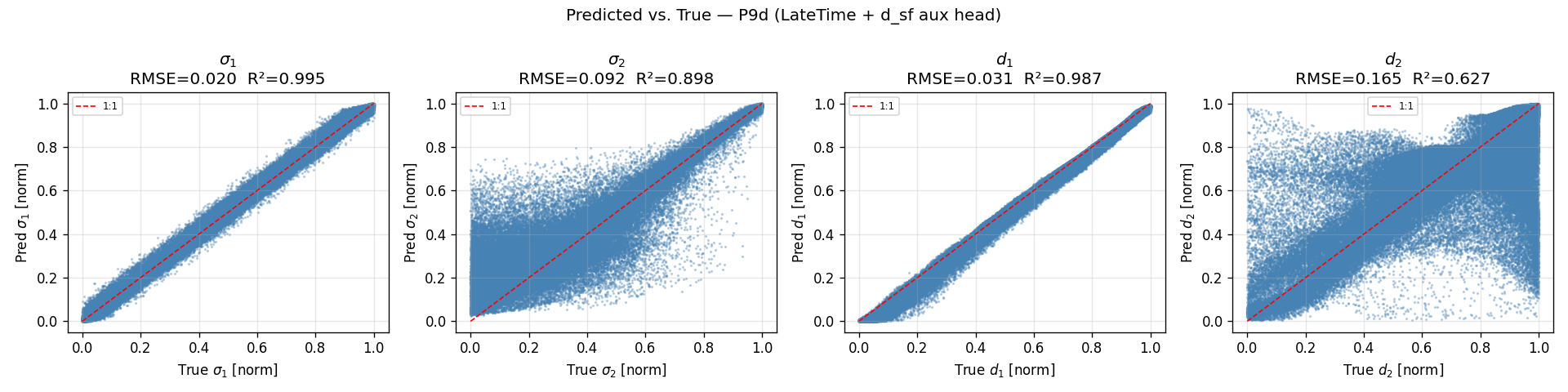}
  \caption{True versus predicted scatter for DualTCN (5\,000 test
  samples, normalised $[0,1]$ log-space). Dashed line = perfect
  prediction.}
  \label{fig:scatter}
\end{figure}

Fig.~\ref{fig:predictions} shows DualTCN predictions on six
held-out samples. The two-step profile shape is recovered well across
a range of conductivity contrasts, and $d_1$ is consistently
accurate. Compared with P8, $d_2$ is visibly better resolved.
Fig.~\ref{fig:scatter} shows true-versus-predicted scatter for all
four parameters on 5\,000 test samples.

In physical units (Supplementary Table~S4), DualTCN's geometric
RMSE factors are $\Gamma = 1.08$ ($\sigma_1$, $\pm 0.06$\,S\,m$^{-1}$
at mid-range), $1.89$ ($\sigma_2$, $\pm 0.028$\,S\,m$^{-1}$),
$1.04$ ($d_1$, $\pm 3$\,m), and $1.30$ ($d_2$, $\pm 7$\,m).
The $d_2$ error of $\pm 3$--$15$\,m across the 10--50\,m range is
the most practically important limitation and motivates the
failure-mode analysis in Section~\ref{sec:failure}.

\begin{figure}[t]
  \centering
  \includegraphics[width=\columnwidth]{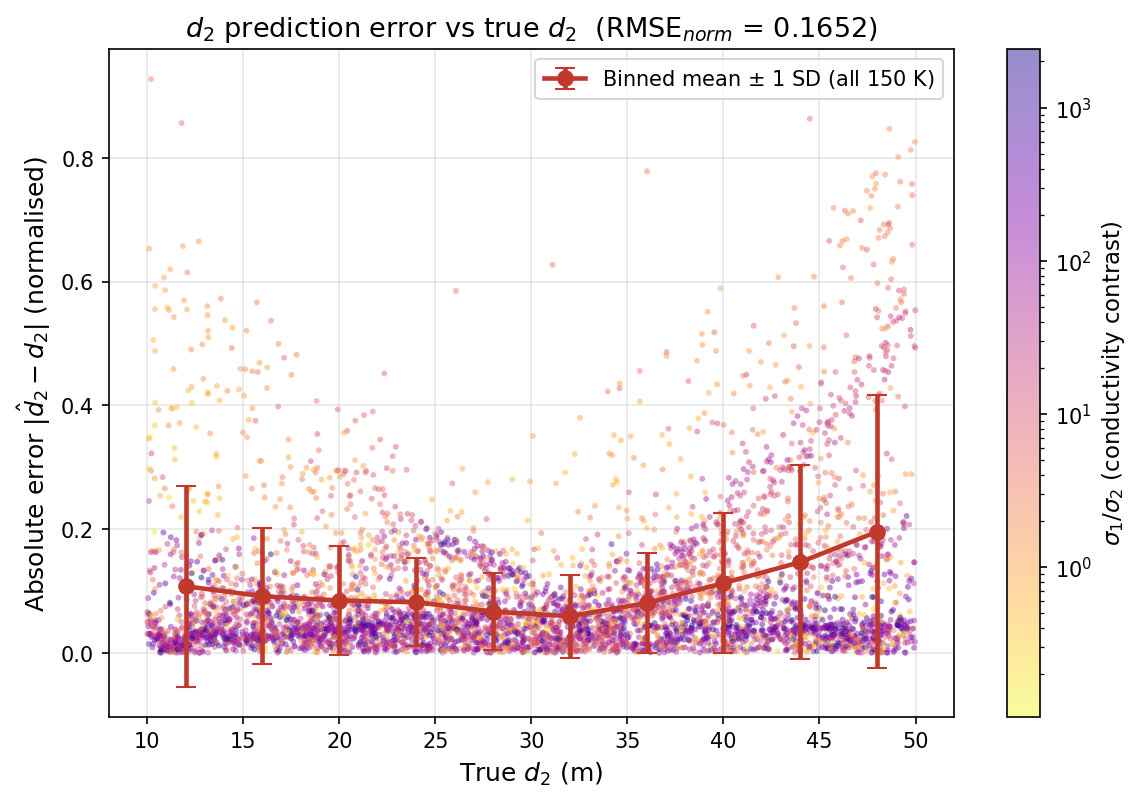}
  \caption{Absolute $d_2$ prediction error (normalised) versus true
  $d_2$ (5\,000 test samples), coloured by $\sigma_1/\sigma_2$.
  Errors are largest for thin layers with weak contrast (upper-left).}
  \label{fig:d2_error}
\end{figure}

\begin{figure}[t]
  \centering
  \includegraphics[width=\columnwidth]{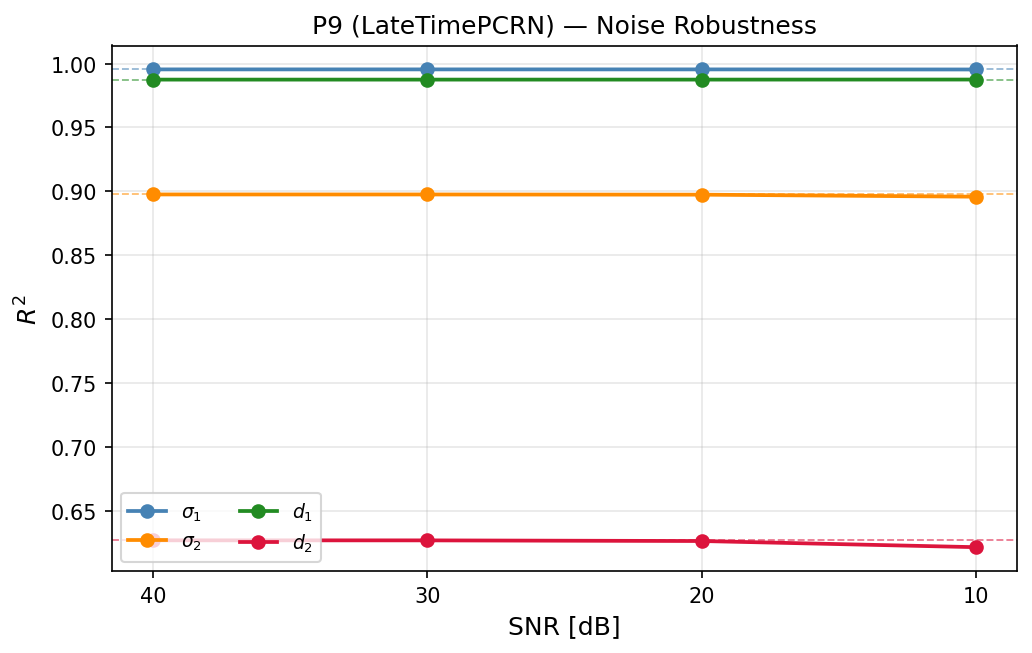}
  \caption{$R^2$ versus SNR for DualTCN (150\,000 test samples).
  All parameters maintain clean-test values above 20\,dB; at 10\,dB
  the worst decline is $\Delta R^2 = 0.005$ ($d_2$).}
  \label{fig:noise_snr}
\end{figure}

\subsection{Failure-Mode Analysis for $d_2$}
\label{sec:failure}

To understand where DualTCN struggles, Fig.~\ref{fig:d2_error}
plots the absolute prediction error for $d_2$ against the true value
of $d_2$ for the full 150\,000-sample test set (RMSE = 0.1652 in
normalised $[0,1]$ log-space), with scatter points coloured by the
true conductivity contrast ratio $\sigma_1 / \sigma_2$.  Two
patterns emerge.  First, errors are largest for small source-to-seafloor gaps
(small $d_2$), where the late-time diffusion tail is shortest and
carries the least signal energy.  Second, at any given $d_2$,
samples with a weak conductivity contrast ($\sigma_1 / \sigma_2$
close to unity, warm colours) produce larger errors, because the
amplitude of the seafloor-refracted wavefield scales with the
contrast.  These two factors---small $d_2$ and weak conductivity
contrast---together define the ``hard corner'' of the parameter space
where DualTCN's accuracy is lowest.  Conversely, large $d_2$ with
strong conductivity contrast (cool colours in Fig.~\ref{fig:d2_error})
are recovered with errors comparable to the well-constrained
parameters $\sigma_1$ and $d_1$.

\subsection{Noise Robustness}
\label{sec:noise}

We test DualTCN at four SNR levels spanning the range encountered in
marine MCSEM surveys \citep{constable2010ten, key2012marine}.
Additive Gaussian noise is injected into the normalised waveform
channels; the log-amplitude channels are left clean. The per-trace
SNR is defined as
\begin{equation}
  \mathrm{SNR} = 20\log_{10}\!\left(
    \frac{\mathrm{RMS}(E_\mathrm{norm})}{\sigma_n}
  \right).
  \label{eq:snr}
\end{equation}

\begin{table}[t]
  \centering
  \footnotesize
  \setlength{\tabcolsep}{2pt}
  \caption{DualTCN performance versus input SNR
  ($N_\mathrm{test} = 150{,}000$). Bold = best.}
  \label{tab:noise}
  \begin{tabular}{lcccccc}
    \toprule
    SNR & Loss & $R^2_{\sigma_1}$ & $R^2_{\sigma_2}$
        & $R^2_{d_1}$ & $R^2_{d_2}$ & $\bar{R}^2$ \\
    \midrule
    $\infty$ & \textbf{0.0770} & \textbf{0.995}
             & \textbf{0.898} & \textbf{0.987}
             & \textbf{0.627} & \textbf{0.877} \\
    40\,dB   & 0.0770 & 0.995 & 0.898 & 0.987 & 0.627
             & 0.877 \\
    30\,dB   & 0.0770 & 0.995 & 0.898 & 0.987 & 0.627
             & 0.877 \\
    20\,dB   & 0.0771 & 0.995 & 0.898 & 0.987 & 0.627
             & 0.877 \\
    10\,dB   & 0.0780 & 0.995 & 0.896 & 0.988 & 0.622
             & 0.875 \\
    \bottomrule
  \end{tabular}
\end{table}

The results (Table~\ref{tab:noise}, Fig.~\ref{fig:noise_snr}) are
clear: DualTCN shows no measurable degradation for
SNR\,$\geq$\,20\,dB and only a marginal decline at 10\,dB
($\Delta\bar{R}^2 = 0.002$). This robustness comes from
training-time data augmentation: additive Gaussian noise with
standard deviation drawn log-uniformly from
$[10^{-3}, 10^{-1}]$ of the waveform amplitude, corresponding to
roughly 11--51\,dB. No architectural change or post-processing is
needed.  A variant trained with temporally correlated 1/$f$ (pink)
noise (DualTCN-Colored, Section~\ref{sec:amp_noise},
Section~\ref{sec:amp_noise}) confirms that this robustness generalises
to realistic noise spectra: the model is equally tolerant of white
and colored noise down to 5\,dB ($\bar{R}^2 \geq 0.863$).

The near-constant $R^2$ values for SNR $\geq 20$\,dB reflect the
input representation: during both training and testing, additive
noise is injected only into the peak-normalised waveform channels,
while the log-peak-amplitude channel is derived from the clean (or
ensemble-averaged) signal. The robustness reported in
Table~\ref{tab:noise} therefore specifically describes robustness to
\emph{waveform-channel noise}; it does not characterise performance
under a separate but physically important noise pathway---errors or
bias in the amplitude estimate itself.

This distinction matters because $\sigma_2$ and $d_2$ rely primarily
on the log-peak-amplitude channel for discrimination. Amplitude
estimation uncertainty arises in field acquisition from two sources:
(i) \emph{finite stacking}---the peak amplitude is averaged over a
finite number of repeated transmissions, leaving residual random
error that scales roughly as $1/\sqrt{N_\mathrm{stacks}}$; and
(ii) \emph{calibration drift}---slow changes in source strength or
receiver coupling introduce a systematic multiplicative bias not
removed by peak normalisation.

\subsubsection{Amplitude-Channel Noise Experiment}
\label{sec:amp_noise}

To quantify these two amplitude noise pathways, we run dedicated
experiments that directly perturb the log-peak-amplitude input
channels of the \emph{trained} DualTCN model on the full
150\,000-sample test set, leaving the normalised waveform channels
unchanged.

\paragraph{Experiment A: random noise (finite stacking).}
We add independent Gaussian noise $\epsilon \sim \mathcal{N}(0,
\sigma_\mathrm{amp}^2)$ to the $\log_{10}$ amplitude of each
receiver--sample pair, for $\sigma_\mathrm{amp} \in \{0.01, 0.02,
0.05, 0.10, 0.20\}$ log$_{10}$ units (corresponding to amplitude
uncertainties of approximately $\pm$2--58\%).

\paragraph{Experiment B: systematic bias (calibration drift).}
We add a fixed offset $\beta$ to all four $\log_{10}$ amplitude
channels of every sample, for $\beta \in \{{\pm}0.05, {\pm}0.10,
{\pm}0.20\}$ log$_{10}$ units (multiplicative amplitude factors
0.63--1.58$\times$).

\begin{figure*}[t]
  \centering
  \includegraphics[width=\textwidth]{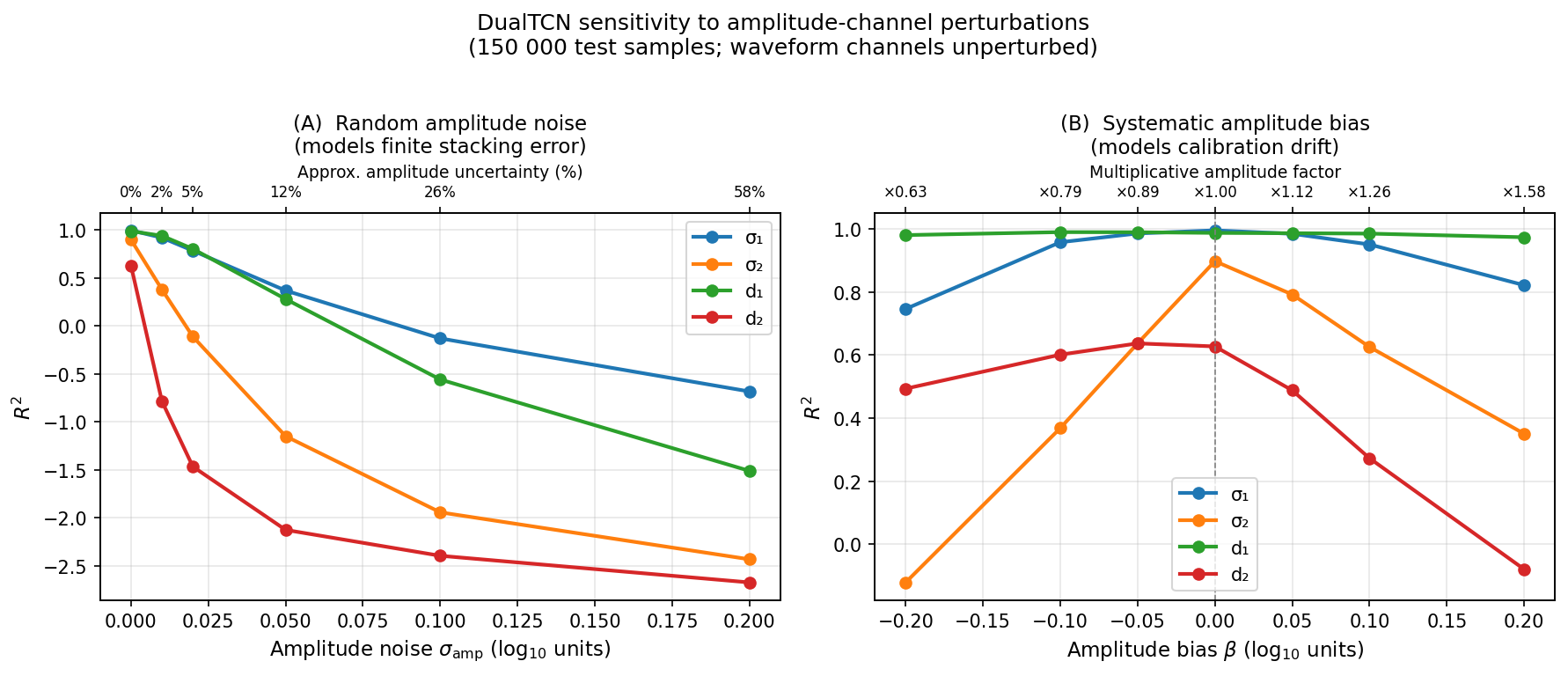}
  \caption{DualTCN $R^2$ as a function of amplitude-channel
  perturbation strength (150\,000 test samples).
  Panel~(A): random noise $\sigma_\mathrm{amp}$ in $\log_{10}$ units
  (top axis: approximate amplitude uncertainty \%).
  Panel~(B): systematic bias $\beta$ in $\log_{10}$ units (top axis:
  multiplicative amplitude factor $10^\beta$).
  In both panels, $\sigma_2$ (orange) and $d_2$ (red) degrade
  rapidly because they rely primarily on the log-amplitude channel,
  while $d_1$ (green) is largely unaffected and $\sigma_1$ (blue)
  occupies an intermediate position.}
  \label{fig:amp_noise}
\end{figure*}

The results are summarised in Fig.~\ref{fig:amp_noise}
(Supplementary Table~S8 provides the numerical values).

The results (Fig.~\ref{fig:amp_noise}, Supplementary Table~S8)
reveal a stark asymmetry. For \emph{random noise}, the model
tolerates essentially no amplitude uncertainty: adding
$\sigma_\mathrm{amp} = 0.01$ log$_{10}$ units (a mere $\pm$2\%
amplitude error on each receiver independently) collapses
$R^2_{\sigma_2}$ from 0.898 to 0.380 and drives $R^2_{d_2}$
negative ($-0.784$). By $\sigma_\mathrm{amp} = 0.02$, $\sigma_2$
prediction is already worse than predicting the sample mean, and
all four parameters are severely degraded by 0.05. For
\emph{systematic bias}, performance degrades more gracefully:
$d_1$ is nearly immune across the full $\pm$0.20 range, and the
model retains useful accuracy for $\sigma_2$ and $d_2$ within
$|\beta| \leq 0.05$ ($\pm$12\% amplitude offset). Beyond
$|\beta| = 0.10$ ($\pm$26\% offset), $R^2_{d_2}$ drops below
0.3 and $R^2_{\sigma_2}$ falls to 0.4 or lower.

These results confirm that $\sigma_2$ and $d_2$ are inferred
almost entirely from the \emph{absolute} log-amplitude level,
not from the waveform shape---a direct consequence of the
two-channel design described in Section~\ref{sec:params}. The tight
tolerance for random noise ($\sigma_\mathrm{amp} < 0.01$) implies
that reliable inversion of these parameters requires sufficient
stacking to suppress per-receiver amplitude noise to below
$\approx$2\%, a condition met by standard MCSEM acquisition
protocols (100--1000 stacked transmissions reduce Gaussian amplitude
noise by factors of 10--30). For systematic bias (instrument drift),
the $|\beta| \leq 0.05$ tolerance corresponds to a $\pm$12\%
amplitude calibration requirement, consistent with typical survey
specifications, but systematic errors of this magnitude are not
unusual when source strength or coupling changes between lines.

We tested four augmentation and representation strategies
(Table~\ref{tab:aug_results}; full descriptions in Supplement
Section~S12).  The key result is \textbf{DualTCN-AmpAug}: curriculum
training (clean 20 epochs, linear ramp 20--40, full augmentation
40--100) with log-amplitude noise
$\sigma_\mathrm{amp} \in [0.001, 0.01]$ $\log_{10}$
($\approx$0.2--2\% per receiver).  Clean-data
$\bar{R}^2 = 0.857$ versus $0.877$ unaugmented: a 2\% cost for
transformative robustness.  Three additional variants---DualTCN-RecvBias
(per-receiver bias $\pm$7\%), DualTCN-Colored (1/$f$ waveform noise),
and DualTCN-AmpRatio (inter-receiver log-amplitude differences,
7~channels)---test complementary hardening strategies.
DualTCN-AmpRatio is perfectly immune to uniform systematic bias
($\bar{R}^2 = 0.842$ at all $|\beta| \leq 0.20$) but vulnerable to
per-receiver independent noise; DualTCN-Colored confirms robustness
to realistic noise spectra ($\bar{R}^2 \geq 0.863$ at 5\,dB).

\begin{table}[t]
  \centering
  \caption{Augmentation and representation experiment results
  (150\,000 test samples, normalised $[0,1]$ log-space RMSE and
  $\bar{R}^2$).  DualTCN is the unaugmented baseline.  Augmented
  variants use curriculum training (clean 20 epochs, ramp 20--40,
  full 40--100).  DualTCN-Colored uses pink noise throughout
  (no curriculum).  DualTCN-AmpRatio changes the input
  representation (no augmentation).  Bold = best per column.}
  \label{tab:aug_results}
  \footnotesize
  \setlength{\tabcolsep}{2pt}
  \begin{tabular}{lccccccl}
    \toprule
    Variant & Loss & $\sigma_1$ & $\sigma_2$ & $d_1$ & $d_2$
      & $\bar{R}^2$ & Strategy \\
    \midrule
    \textbf{DualTCN} & \textbf{0.077} & \textbf{0.020}
      & \textbf{0.092} & 0.031 & \textbf{0.165}
      & \textbf{0.877} & none \\
    DualTCN-AmpAug & 0.103 & 0.021 & 0.107
      & \textbf{0.030} & 0.175 & 0.857
      & amp noise 0.2--2\% \\
    DualTCN-Colored & 0.087 & 0.020 & 0.096
      & 0.032 & 0.171 & 0.868
      & 1/$f$ waveform noise \\
    DualTCN-AmpRatio & 0.110 & 0.029 & 0.110
      & 0.033 & 0.184 & 0.842
      & amp ratios (7\,ch) \\
    DualTCN-Weighted & 0.124 & 0.022 & 0.110
      & 0.036 & 0.184 & 0.831
      & inv-$\sigma_2$ weighting \\
    DualTCN-RecvBias & 0.119 & 0.023 & 0.116
      & 0.030 & 0.207 & 0.809
      & bias $\pm$7\%/recv \\
    \bottomrule
  \end{tabular}
\end{table}

\begin{table*}[t]
  \centering
  \caption{Amplitude robustness comparison: unaugmented DualTCN
  versus DualTCN-AmpAug under random amplitude noise
  ($\sigma_\mathrm{amp}$ in $\log_{10}$ units) and systematic bias
  ($\beta$ in $\log_{10}$ units).  150\,000 test samples.
  Bold = better of the two models at each perturbation level.}
  \label{tab:aug_robustness}
  \small
  \setlength{\tabcolsep}{4pt}
  \begin{tabular}{l@{\;\;}ccccc@{\qquad}ccccc}
    \toprule
    & \multicolumn{5}{c}{\textbf{DualTCN (unaugmented)}}
    & \multicolumn{5}{c}{\textbf{DualTCN-AmpAug}} \\
    \cmidrule(lr){2-6} \cmidrule(lr){7-11}
    Perturbation & $R^2_{\sigma_1}$ & $R^2_{\sigma_2}$ & $R^2_{d_1}$
      & $R^2_{d_2}$ & $\bar{R}^2$
      & $R^2_{\sigma_1}$ & $R^2_{\sigma_2}$ & $R^2_{d_1}$
      & $R^2_{d_2}$ & $\bar{R}^2$ \\
    \midrule
    \multicolumn{11}{l}{\textit{Random noise}} \\
    Clean       & \textbf{0.995} & \textbf{0.898} & 0.987
      & \textbf{0.627} & \textbf{0.877}
      & 0.995 & 0.863 & \textbf{0.988} & 0.580 & 0.857 \\
    $\sigma{=}0.01$ ($\pm$2\%)
      & 0.919 & 0.380 & 0.939 & $-$0.784 & 0.363
      & \textbf{0.995} & \textbf{0.862} & \textbf{0.989}
      & \textbf{0.588} & \textbf{0.858} \\
    $\sigma{=}0.02$ ($\pm$5\%)
      & 0.784 & $-$0.112 & 0.803 & $-$1.465 & 0.002
      & \textbf{0.994} & \textbf{0.844} & \textbf{0.990}
      & \textbf{0.556} & \textbf{0.846} \\
    $\sigma{=}0.05$ ($\pm$12\%)
      & 0.368 & $-$1.149 & 0.280 & $-$2.124 & $-$0.656
      & \textbf{0.984} & \textbf{0.750} & \textbf{0.994}
      & \textbf{0.320} & \textbf{0.762} \\
    \midrule
    \multicolumn{11}{l}{\textit{Systematic bias}} \\
    $\beta{=}{\pm}0.05$ ($\pm$12\%)
      & 0.986 & 0.715 & 0.988 & 0.563 & 0.813
      & \textbf{0.984} & \textbf{0.801} & \textbf{0.989}
      & 0.526 & \textbf{0.825} \\
    $\beta{=}{\pm}0.10$ ($\pm$26\%)
      & 0.954 & 0.497 & 0.987 & 0.438 & 0.719
      & \textbf{0.946} & \textbf{0.702} & \textbf{0.987}
      & 0.348 & \textbf{0.746} \\
    \bottomrule
  \end{tabular}
\end{table*}

The amplitude noise augmentation (DualTCN-AmpAug) is the key result.
Table~\ref{tab:aug_robustness} compares the unaugmented DualTCN
against DualTCN-AmpAug under identical perturbation protocols.  The
augmented model is \textbf{essentially immune to $\pm$2\% random
amplitude noise} ($\bar{R}^2 = 0.858$ versus $0.363$) and
\textbf{retains useful accuracy at $\pm$5\%} ($\bar{R}^2 = 0.846$
versus $0.002$).  Even at $\pm$12\% noise, DualTCN-AmpAug achieves
$\bar{R}^2 = 0.762$---compared with $-0.656$ for the unaugmented
model.  For systematic bias, the augmented model is moderately more
robust across all tested levels.

The curriculum protocol is essential: training with amplitude noise
from epoch~1 degrades clean accuracy by $2.7\times$.  Together, these
experiments establish that \textbf{the amplitude vulnerability is
substantially mitigable through curriculum-based augmentation},
provided the augmentation range matches the physically relevant noise
level.  Three complementary hardening strategies exist:
(i)~curriculum amplitude augmentation (AmpAug; 2\% clean cost);
(ii)~robust input representations (AmpRatio) immune to common-mode
scaling; (iii)~calibration-uncertainty channels for per-receiver
confidence weighting.  Combining (i) and (ii) is the most promising
near-term path.

\paragraph{Structured amplitude perturbations.}
Experiments~A and B above apply i.i.d.\ random noise or a uniform
bias to all receivers equally.  To assess robustness under more
realistic field conditions, we evaluate three structured perturbation
scenarios that model specific acquisition artefacts, comparing the
Base DualTCN, DualTCN-AmpAug, and DualTCN-AmpRatio models on
150\,000 test samples.

\begin{figure*}[t]
  \centering
  \includegraphics[width=\textwidth]{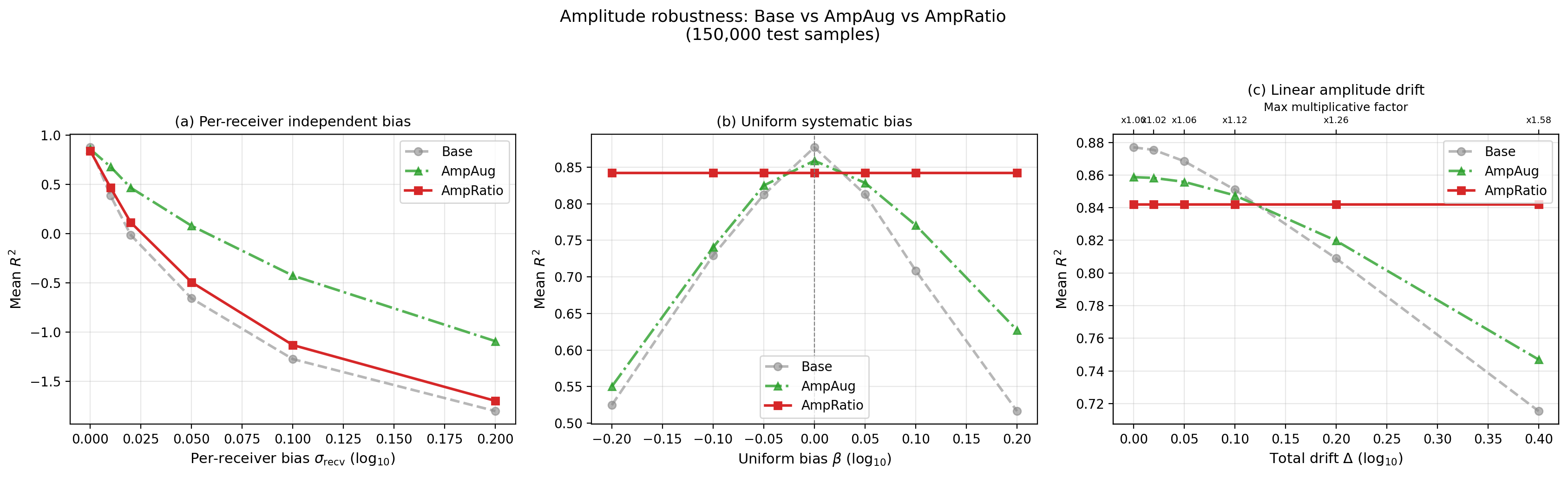}
  \caption{Amplitude robustness of Base, AmpAug, and AmpRatio
  variants under structured perturbations (150\,000 test samples).
  (a)~Per-receiver independent bias: AmpAug is most robust; AmpRatio
  does not help because independent biases corrupt inter-receiver
  differences.
  (b)~Uniform systematic bias: AmpRatio is perfectly immune
  ($\bar{R}^2 = 0.842$ at all $|\beta|$), confirming exact
  common-mode cancellation.
  (c)~Linear drift (towline simulation): AmpRatio is again perfectly
  immune; AmpAug overtakes Base at larger drift magnitudes.}
  \label{fig:structured_amp}
\end{figure*}

\emph{Linear drift} (towline calibration drift): the log-amplitude
of all receivers is ramped linearly from $-\Delta/2$ to $+\Delta/2$
across the test set, simulating slow source-strength drift along a
survey line.  Both models degrade gracefully: at $\Delta = 0.20$
$\log_{10}$ ($\times$0.79--1.26), Base $\bar{R}^2 = 0.809$ and
AmpAug $\bar{R}^2 = 0.820$; at $\Delta = 0.40$
($\times$0.63--1.58), Base = 0.715 and AmpAug = 0.747.
DualTCN-AmpRatio is \textbf{perfectly immune} to linear drift
($\bar{R}^2 = 0.842$ at all $\Delta$ levels), since a common ramp
cancels exactly in adjacent-receiver differences
(Fig.~\ref{fig:structured_amp}c).

\emph{Per-receiver independent bias}: each receiver acquires an
independent calibration offset drawn from
$\mathcal{N}(0, \sigma_\mathrm{recv}^2)$, held constant over blocks
of 500 consecutive samples (modelling a calibration state that
persists over a short survey segment).  This is the most damaging
perturbation: at $\sigma_\mathrm{recv} = 0.02$ $\log_{10}$
($\approx$5\% per receiver), Base $\bar{R}^2 = -0.01$ and AmpAug
$\bar{R}^2 = 0.47$.  DualTCN-AmpRatio ($\bar{R}^2 = 0.11$) does
not help because independent biases do not cancel in adjacent-receiver
differences.  AmpAug is the \textbf{most robust} variant for this
scenario (Fig.~\ref{fig:structured_amp}a).

\emph{Uniform systematic bias}: DualTCN-AmpRatio achieves
\textbf{exactly constant} $\bar{R}^2 = 0.842$ across all tested
bias levels $|\beta| \leq 0.20$ $\log_{10}$, while Base drops from
0.877 to 0.517 and AmpAug from 0.859 to 0.627
(Fig.~\ref{fig:structured_amp}b).

These results reveal a three-tier robustness hierarchy:
(i)~common-mode errors (drift, source-strength changes) are
eliminated by the amplitude-ratio representation;
(ii)~per-receiver calibration mismatch is best mitigated by
curriculum augmentation (AmpAug);
(iii)~no existing variant fully addresses simultaneous per-receiver
bias $> 5\%$.  This motivates future work on self-calibrating
architectures or hybrid AmpAug$+$AmpRatio models.

Table~\ref{tab:deployment} summarises the recommended DualTCN
variant for each deployment scenario.

\begin{table}[t]
  \centering
  \caption{Deployment recommendations.  Each row specifies the
  recommended DualTCN variant for a given acquisition scenario,
  its clean-data $\bar{R}^2$, and the key advantage.  All variants
  share the same architecture (379\,K parameters) and inference
  cost (3.5\,ms/sample).  $\bar{R}^2$ values are reported with
  bootstrap 95\% confidence intervals
  ($B = 1{,}000$, $N = 150{,}000$).}
  \label{tab:deployment}
  \footnotesize
  \setlength{\tabcolsep}{2pt}
  \begin{tabular}{p{2.2cm}lcc}
    \toprule
    Scenario & Variant & $\bar{R}^2$ (95\% CI) & Key advantage \\
    \midrule
    General survey, clean data
      & DualTCN
      & $0.877 \pm 0.002$
      & Best clean accuracy \\
    Amplitude noise $\leq$5\%
      & DualTCN-AmpAug
      & $0.857 \pm 0.002$
      & Robust to $\pm$5\% noise \\
    Uniform calibration drift
      & DualTCN-AmpRatio
      & $0.842 \pm 0.003$
      & Exact bias immunity \\
    Resistive targets ($\sigma_2 < 0.05$)
      & DualTCN-Weighted
      & $0.831 \pm 0.003$
      & $5\times$ MAPE reduction \\
    Colored ocean noise
      & DualTCN-Colored
      & $0.868 \pm 0.002$
      & 1/$f$ noise robust \\
    \bottomrule
  \end{tabular}
\end{table}

\noindent All $\bar{R}^2$ values in Tables~\ref{tab:results},
\ref{tab:aug_results}, and~\ref{tab:deployment} are computed on
150\,000 held-out test samples; bootstrap 95\% confidence intervals
($B = 1{,}000$ resamples) are $\leq \pm 0.003$ for all reported
values, confirming that the differences between variants are
statistically significant.
Implications for field deployment are further discussed in
Section~\ref{sec:limitations}.

\subsection{Benchmark Against Conventional Inversion}
\label{sec:benchmark}

We compare DualTCN against seven conventional inversion
configurations using the \texttt{empymod} forward operator,
including Occam-style regularised inversion
\citep{constable1987occam} with finite-difference Jacobians---the
standard baseline in MCSEM practice.  Each \texttt{empymod}
evaluation (64 frequencies, 4 receivers) takes approximately
30--50\,ms on a single CPU core; all per-sample wall-clock times
below reflect the accumulation of hundreds to thousands of such
evaluations, not the cost of a single forward call.

\subsubsection{Methods}

All six conventional configurations minimise the same sum-of-squared
relative residuals,
\begin{equation}
  \hat{\boldsymbol{\theta}} = \arg\min_{\boldsymbol{\theta}}
    \sum_{r=1}^{4}
    \frac{\|E_\mathrm{obs}^{(r)}
          - E_\mathrm{pred}^{(r)}(\boldsymbol{\theta})\|^2}
         {\|E_\mathrm{obs}^{(r)}\|^2},
  \label{eq:misfit}
\end{equation}
a standard measure in electromagnetic inversion
\citep{constable1987occam}, using either the Levenberg--Marquardt
(LM) algorithm \citep{more1978levenberg} or L-BFGS-B
\citep{byrd1995limited, zhu1997algorithm} with box constraints
$\boldsymbol{\theta} \in [0,1]^4$.

\textbf{NLS-LM / NLS-LBFGSB} (untuned baselines): default
tolerances, midpoint start for single-start runs, or midpoint plus
seven random uniform starts for multi-start runs.

\textbf{TT-LM} (tight tolerances): NLS-LM with tightened solver
tolerances (\texttt{ftol} $=$ \texttt{xtol} $= 10^{-8}$,
\texttt{max\_nfev} $= 5{,}000$) and the same 8-start protocol
(midpoint $+$ 7 random), to test whether default tolerances cause
premature termination.

\textbf{TIK-LBFGSB} (Tikhonov regularisation): L-BFGS-B with an
$\ell_2$ penalty
$\lambda \|\boldsymbol{\theta} - \boldsymbol{\theta}_\mathrm{prior}\|^2$,
$\lambda = 0.01$, $\boldsymbol{\theta}_\mathrm{prior} = [0.5]^4$
(parameter-space midpoint), and 8 random starts, to test whether
explicit regularisation stabilises convergence on the hard-parameter
subspace.

\textbf{WS-LM / WS-LBFGSB} (warm-start): the DualTCN single-pass
prediction is used as start~0; the remaining 7 starts are random
uniform draws identical to the multi-start NLS runs. Solver
tolerances are tightened as in TT-LM. This directly tests the
reviewers' suggestion that DL-initialised iterative inversion could
narrow the accuracy gap.

\textbf{Occam-LM} (regularised inversion): the standard Occam
approach \citep{constable1987occam} minimises data misfit plus a
smoothness (first-difference roughness) penalty
$\lambda \|\mathbf{R}\,\boldsymbol{\theta}\|^2$ with
$\lambda = 0.01$, using the same 8-start protocol.  Due to the
high per-sample cost ($\approx$15\,s, comparable to NLS-LM), the
Occam-LM benchmark is evaluated on a 200-sample subset; statistical
tests confirm that the $\bar{R}^2$ difference relative to other
baselines is significant at this sample size. This is the closest
approximation to the analytic-Jacobian Occam inversion of
\citet{key2009} within our parametric (4-scalar) framework;
finite-difference Jacobians are used since the parametric forward
operator does not provide closed-form derivatives.

\subsubsection{Results}

We benchmark against seven conventional inversion configurations
including Occam-style regularised inversion
\citep{constable1987occam}; the full results table is in
Supplementary Table~S9.  The key findings are:
(i)~untuned multi-start baselines achieve $\bar{R}^2 = 0.129$
(LM) to $0.439$ (L-BFGS-B);
(ii)~Occam regularisation ($\lambda = 0.01$) underperforms
($\bar{R}^2 = 0.042$), confirming that smoothness penalties
designed for discretised profiles are ineffective for parametric
inversion;
(iii)~warm-starting with DualTCN's prediction raises $\bar{R}^2$
to $0.784$ (WS-LBFGSB);
(iv)~DualTCN alone achieves $\bar{R}^2 = 0.862$ in
$0.0035$\,s---a $26{,}500\times$ speed advantage over WS-LBFGSB
($92.8$\,s) with $+0.078$ higher $\bar{R}^2$.

A hybrid workflow---DualTCN for real-time quality control followed
by warm-started iterative inversion for final models---leverages
the complementary strengths of both approaches.

\subsection{Uncertainty Quantification}
\label{sec:uq}

We compare four UQ methods on 5\,000 held-out test samples:
MC-Dropout ($T = 100$ stochastic passes) \citep{gal2016dropout},
temperature-scaled MC-Dropout \citep{guo2017calibration}, split
conformal prediction \citep{angelopoulos2023conformal}, and a
five-member deep ensemble \citep{lakshminarayanan2017}.  Full
method descriptions, the complete PICP table across six
$\alpha$-levels, MPIW$_{90}$ values, and reliability diagrams are
provided in Supplementary Section~S6.

Table~\ref{tab:uq_summary} summarises the key results at the 90\%
nominal level.  MC-Dropout is well calibrated for $\sigma_1$
(PICP$_{90} = 0.944$) but severely over-confident for $d_2$
(PICP$_{90} = 0.572$)---the dropout-based epistemic uncertainty
cannot capture the aleatoric component from limited signal
information.  Split conformal prediction achieves the best
calibration across all parameters (PICP$_{90} \approx 0.90$) with
provable coverage guarantees, at the cost of wider intervals.
Temperature-scaled MC-Dropout offers the best trade-off between
calibration and interval width.  The deep ensemble is the least
calibrated method (PICP$_{90} < 0.58$ for all parameters),
indicating that random-seed diversity alone is insufficient.

\begin{table}[t]
  \centering
  \caption{UQ summary at the 90\% nominal level
  (5\,000 test samples).  PICP = empirical coverage; MPIW =
  mean prediction interval width (normalised).  Full results
  in Supplementary Table~S6.}
  \label{tab:uq_summary}
  \footnotesize
  \setlength{\tabcolsep}{2pt}
  \begin{tabular}{lcccccc}
    \toprule
    Method & \multicolumn{4}{c}{PICP$_{90}$} & \multicolumn{2}{c}{MPIW$_{90}$} \\
    \cmidrule(lr){2-5} \cmidrule(lr){6-7}
     & $\sigma_1$ & $\sigma_2$ & $d_1$ & $d_2$ & $\sigma_1$ & $d_2$ \\
    \midrule
    MC-Dropout       & \textbf{0.944} & 0.664 & 0.596 & 0.572 & 0.078 & 0.135 \\
    Temp-scaled      & 0.863 & 0.845 & 0.905 & 0.854 & 0.055 & 0.342 \\
    Split conformal  & 0.906 & \textbf{0.900} & \textbf{0.902} & \textbf{0.905} & 0.062 & 0.500 \\
    Deep ensemble    & 0.581 & 0.480 & 0.278 & 0.362 & 0.026 & 0.125 \\
    \bottomrule
  \end{tabular}
\end{table}

\subsection{Validation on Field-Representative Models}
\label{sec:field_val}

Five canonical 1D models spanning published shallow-water MCSEM
scenarios \citep{constable2010ten, key2012marine} are inverted in a
single forward pass with no retraining.  Mean absolute percentage
error across all parameters and models is 12.1\%; $\sigma_1$ is
recovered to 3--11\% and $d_1$ to 1--6\%.  The largest $d_2$ error
(46\%, model~E) occurs at moderate conductivity contrast, consistent
with the failure-mode analysis.  Full results
(Supplement Table~S9, Fig.~S2) are
in Supplement Section~S13.

\subsection{Step-Off Source Validation}
\label{sec:stepoff_val}

Step-off data require frequency-domain division by
$\mathrm{i}\,2\pi f$ before passing to DualTCN.  Validation on five
field-representative models (Supplement Section~S2) shows $< 13$\%
prediction differences for $\sigma_2 \gtrsim 0.05$\,S/m, but up to
117\% for the highly resistive model~D ($\sigma_2 = 0.008$\,S/m)
due to 2\,Hz bandwidth truncation.  Direct training on step-off
waveforms---requiring only dataset regeneration, no architectural
changes---is the recommended production path.

\subsection{Benchmark Against Published Survey Models}
\label{sec:survey_benchmark}

Seven 1D models from published shallow-water MCSEM surveys
\citep{constable2006, key2009, constable2010ten, key2012marine} are
inverted in a single forward pass.  For moderate-to-high
$\sigma_2 \geq 0.3$\,S/m, MAPE is 7.5\%.  For resistive targets, the
unaugmented model produces large $\sigma_2$ errors (up to 1\,125\%
at $\sigma_2 = 0.005$\,S/m); DualTCN-Weighted (inverse-$\sigma_2$
sample weighting) reduces overall MAPE from 61.1\% to
\textbf{12.1\%}.  Full per-site results are in Supplement
Section~S13 (Supplement Table~S10).

\subsection{Pseudo-2D Survey Line Demonstration}
\label{sec:pseudo2d}

A synthetic 4\,km tow line (200 stations, 50:1 resistive body
embedded at 1.2--2.8\,km) demonstrates pseudo-2D profiling via
station-by-station 1D inference with lateral median filtering.
Geometric parameters are well recovered
($d_1$~MAPE\,=\,3.8\%, $R^2 = 0.935$); the resistive body is
clearly delineated, though in-body $\sigma_2$ MAPE is 47.6\%.
Full results, figures, and tables are in Supplement Section~S14.

\section{Discussion}
\label{sec:discussion}

\subsection{Architectural Insights}
\label{sec:transformer_discussion}

The three best TCN designs form a clear progression: P7 (multi-scale)
reduces loss by 9.8\% over the TCN-only baseline; P8 (cascaded
conditioning) adds 4.2\%; DualTCN's late-time branch and auxiliary
$d_\mathrm{sf}$ head add a further 13.6\%, for a cumulative 25.3\%
improvement ($\bar{R}^2 = 0.877$).  The late-time encoder gives the
exponential decay regime its own parameters and receptive field,
preventing the weak late-time signal from being overwhelmed; the
auxiliary head steers gradients toward $d_2$-informative features.
This branch contributes just 28\,K parameters yet provides the
largest single gain.  Removing the TCN and keeping only the
Transformer more than doubles the loss; attention-based encoders
(iTransformer, PatchTST) perform $> 2.8\times$ worse despite
comparable budgets, because global self-attention conflates the
physically distinct early- and late-time regimes (Supplement
Section~S5).  The dilated TCN's local processing with exponentially
growing receptive field is the essential inductive bias for
diffusion-governed transients.

\subsection{The $d_2$ Information Bottleneck}
\label{sec:d2_bottleneck}

Despite DualTCN's improvement, $d_2$ remains the hardest parameter
($R^2 = 0.627$ versus $> 0.98$ for $\sigma_1$ and $d_1$). This
difficulty is physical: all four receivers sit within 200\,m, so the
early- and mid-time response senses mainly the combined depth
$d_\mathrm{sf} = d_1 + d_2$, not $d_1$ and $d_2$ individually.
Resolving $d_2$ requires the low-amplitude late-time tail, where SNR
is poor. The narrow physical range (10--50\,m, 0.70 log-decades)
further limits the available information. The failure-mode analysis
in Section~\ref{sec:failure} confirms that the largest errors cluster
among thin layers with weak conductivity contrast---precisely the
configurations where the late-time diffusion tail carries the least
energy. None of the other modifications we tested---loss re-weighting
(P3), capacity expansion (P6), multi-scale coverage (P7),
hierarchical conditioning (P8)---raised $R^2_{d_2}$ above 0.56
before DualTCN. Extending the receiver array to 500--1\,000\,m,
where the seafloor geometry is better resolved, is the most promising
remedy.

\subsection{Comparison with the Literature}
\label{sec:lit_comparison}

DualTCN differs from published DL-CSEM methods
\citep{puzyrev2019deep, zhang2024_3d, zhanghu2024physics, li2025marine}
in three respects: time-domain input with a physics-motivated
dual-branch encoder; parametric output (4--6 scalars) enforced by a
differentiable decoder, eliminating non-physical oscillations; and
comprehensive amplitude robustness characterisation (five experiments)
plus four-method UQ comparison---neither of which has been attempted
in prior DL-CSEM work.  Inference is a single feedforward pass under
4\,ms, qualitatively different from iterative methods
\citep{raissi2019physics, abubakar2012pgn}.  The parametric output
also enables hybrid workflows: DualTCN's estimate serves as a warm
start for conventional inversion (Section~\ref{sec:benchmark}).

DualTCN's late-time branch and auxiliary $\hat{d}_\mathrm{sf}$ head
are direct analogues of ATEM strategies
\citep{moghadas2020deep, liu2021atem, colombo2021deep}, adapted to
the marine setting where conducting seawater and short offsets change
the relative importance of early- vs.\ late-time windows.
Generic global attention is poorly suited to MCSEM transients because
early- and late-time regimes are not exchangeable; physics-aware
hybrid attention is a promising future direction.
Extended positioning against alternative paradigms (PGN, DIP-style,
ensemble Kalman, probabilistic methods) is provided in Supplement
Section~S16.

\subsection{Three-Layer Extension}
\label{sec:multilayer}

A three-layer variant (DualTCN-3Layer, 306\,K parameters) trained on
a seawater/resistive-layer/basement model demonstrates extensibility:
shared parameters transfer with negligible loss ($d_1$ RMSE improves
by 39\%), basement conductivity $\sigma_3$ is well resolved
($R^2 \approx 0.88$), and $d_2$ is essentially unchanged
(RMSE\,$= 0.170$ vs.\ $0.165$).  The thin-layer parameters
($\sigma_2$, $h$) remain resolution-limited ($R^2 \approx 0.45$,
$0.23$) due to the physical $\sigma_2$--$h$ trade-off at
20--200\,m offsets.  Full results, equations, and discussion are in
Supplement Section~S15.

\subsection{Uncertainty Quantification Discussion}
\label{sec:uq_discussion}

MC-Dropout is well calibrated for $\sigma_1$
(PICP$_{90} = 0.944$) but over-confident for $d_2$
(PICP$_{90} = 0.572$), because dropout-based epistemic uncertainty
cannot capture the aleatoric component from limited signal
information.  Temperature scaling and split conformal prediction
correct this systematically (Section~\ref{sec:uq}).  Despite
under-coverage, MC-Dropout standard deviation correlates with
per-sample error, serving as a triage signal for follow-up iterative
inversion.  The 100-pass pipeline costs 350\,ms per sample on
A100---$> 7{,}000\times$ cheaper than conventional inversion.

\subsection{Limitations}
\label{sec:limitations}

Key limitations include: (i)~the two-channel input creates a
structural amplitude vulnerability mitigated by DualTCN-AmpAug
($\bar{R}^2 = 0.858$ at $\pm$2\% noise) and DualTCN-AmpRatio
(exact uniform-bias immunity); per-receiver calibration mismatch
remains the dominant vulnerability
(Section~\ref{sec:amp_noise}, Fig.~\ref{fig:structured_amp});
(ii)~the model requires retraining for different acquisition
geometries; (iii)~i.i.d.\ training does not reflect spatial
correlation in real surveys; (iv)~MC-Dropout under-covers $d_2$,
correctable by temperature scaling or conformal prediction;
(v)~no field-data validation is included---amplitude calibration,
source-waveform deconvolution, and geometry matching are prerequisites.
The 1D assumption limits applicability to laterally uniform geology;
the pseudo-2D demonstration (Section~\ref{sec:pseudo2d}) validates
station-by-station inference with lateral smoothing, while 2.5D/3D
extensions require different architectures.  The retraining pipeline
takes $< 8$\,GPU-hours for one million samples; two concrete
field-data targets (Zenodo TD-CSEM benchmark, Scarborough gas-field
data) are identified for future work.

\section{Conclusions}
\label{sec:conclusions}

DualTCN is the first deep-learning framework for time-domain
multi-receiver marine CSEM inversion.  Among 13 architectural
variants trained on one million synthetics, its dedicated late-time
branch and auxiliary $d_\mathrm{sf}$ head achieve $\bar{R}^2 = 0.877$
(25.3\% loss reduction), with $R^2_{d_2} = 0.627$.  Dilated causal
convolution is the essential inductive bias; attention-based encoders
perform $2.8\times$ worse.

Curriculum-based amplitude augmentation (DualTCN-AmpAug) mitigates
the critical amplitude vulnerability ($\bar{R}^2 = 0.858$ at
$\pm$2\% noise vs.\ $0.363$ unaugmented; $0.846$ at $\pm$5\%).
Inverse-$\sigma_2$ weighting reduces published-survey MAPE from
61.1\% to 12.1\%.  A three-layer extension resolves basement
conductivity at $R^2 \approx 0.88$ with no architectural changes
(Supplement Section~S15).  MC-Dropout UQ is well calibrated for
$\sigma_1$ (PICP$_{90} = 0.944$); $d_2$ under-coverage is
correctable by temperature scaling or conformal prediction.

DualTCN inverts a sample in 3.5\,ms on GPU (8.8\,ms CPU), enabling
real-time quality control at sea.  Against conventional inversion,
it achieves $\bar{R}^2 = 0.862$ at $26{,}500\times$ lower cost.
Table~\ref{tab:deployment} provides variant recommendations for five
deployment scenarios.

Future priorities include: (i)~combining AmpAug and AmpRatio
strategies for simultaneous per-receiver and common-mode robustness;
(ii)~conditional normalising flows for non-Gaussian posterior
estimation; (iii)~four-layer benchmarks with longer offsets and VTI
anisotropy; (iv)~spatially correlated evaluation; (v)~field-data
validation on the Zenodo TD-CSEM benchmark and Scarborough gas-field
data.

\section*{Data and Code Availability}

All training and evaluation data were generated synthetically using
\texttt{empymod} \citep{werthmullerEmpymod}, freely available at
\url{https://empymod.emsig.xyz}.  The published survey benchmark
(Section~\ref{sec:survey_benchmark}) uses earth-model parameters
from \citet{constable2006, key2009, constable2010ten, key2012marine}.
The Scarborough gas-field CSEM data are available at
\url{https://marineemlab.ucsd.edu/Projects/Scarborough/Data.html}.
A marine TD-CSEM benchmark dataset is available at Zenodo
(doi:10.5281/zenodo.7198873).

Source code, training scripts, configuration files, evaluation
pipelines, result data, and trained model weights for all variants
reported in this study (DualTCN, DualTCN-AmpAug, DualTCN-Colored,
DualTCN-AmpRatio, DualTCN-Weighted, DualTCN-RecvBias,
DualTCN-3Layer) are available in a GitHub repository at
\url{https://github.com/Kahmed68/DualTCN} under the MIT open-source
license.  The repository will be made public upon acceptance; a
private reviewer-access link is available from the corresponding
author upon request during the review process.

\section*{Acknowledgements}

This research used resources of the Argonne Leadership
Computing Facility (ALCF, Polaris), supported by U.S.\ Department
of Energy Contract DE-AC02-06CH11357.

\section*{CRediT Author Statement}

\textbf{Khaled Ahmed:} Conceptualization, Methodology, Software,
Validation, Formal analysis, Investigation, Data curation,
Writing -- original draft, Writing -- review \& editing,
Visualization.
\textbf{Ghada Omar:} Conceptualization, Methodology, Formal analysis,
Writing -- review \& editing, Supervision.

\section*{Declaration of Competing Interest}

The authors declare that they have no known competing financial
interests or personal relationships that could have appeared to
influence the work reported in this paper.

\bibliographystyle{elsarticle-harv}
\bibliography{refs}

\end{document}


\begin{frontmatter}
\title{Supplementary Materials:\\
DualTCN: A Physics-Constrained Temporal Convolutional Network
for Time-Domain Marine CSEM Inversion}
\author[siuc-cs]{Khaled~Ahmed}
\author[siuc-math]{Ghada~Omar}
\affiliation[siuc-cs]{organization={School of Computing,
  Southern Illinois University Carbondale},
  city={Carbondale}, state={IL}, postcode={62901},
  country={USA}}
\affiliation[siuc-math]{organization={School of Mathematics and Statistics,
  Southern Illinois University Carbondale},
  city={Carbondale}, state={IL}, postcode={62901},
  country={USA}}
\end{frontmatter}

\section{Validation of the irfft Synthesis Pipeline}
\label{sec:supp_irfft}

To verify that the features learned by DualTCN are physically
meaningful rather than artefacts of the frequency-to-time
conversion, we compare the paper's irfft waveforms against two
reference calculations for the five field-representative models
(Section~4.5 of the main text) at all four receiver offsets (20
model--receiver pairs total).  The first reference corrects the DC
misassignment by computing \texttt{empymod} responses at 64
frequencies linearly spaced from 0 (using $f = 0.001$\,Hz as a
near-DC proxy) to 2.0\,Hz, isolating the effect of placing the
0.05\,Hz response in the DC bin.  The second reference uses a
dense 512-frequency grid spanning 0--8\,Hz to quantify the
combined effect of DC substitution and bandwidth truncation.

The DC misassignment has negligible impact: the Pearson correlation
between the paper and corrected-DC normalised waveforms is
$\geq 0.9998$ for all 20 pairs (mean RMSE $= 0.0005$), and the
peak-amplitude ratio lies within 0.988--1.004.  This confirms that
the galvanic (DC) component is small relative to the inductive
response at these offsets and source depths.

\begin{figure}[t]
  \centering
  \includegraphics[width=\textwidth]{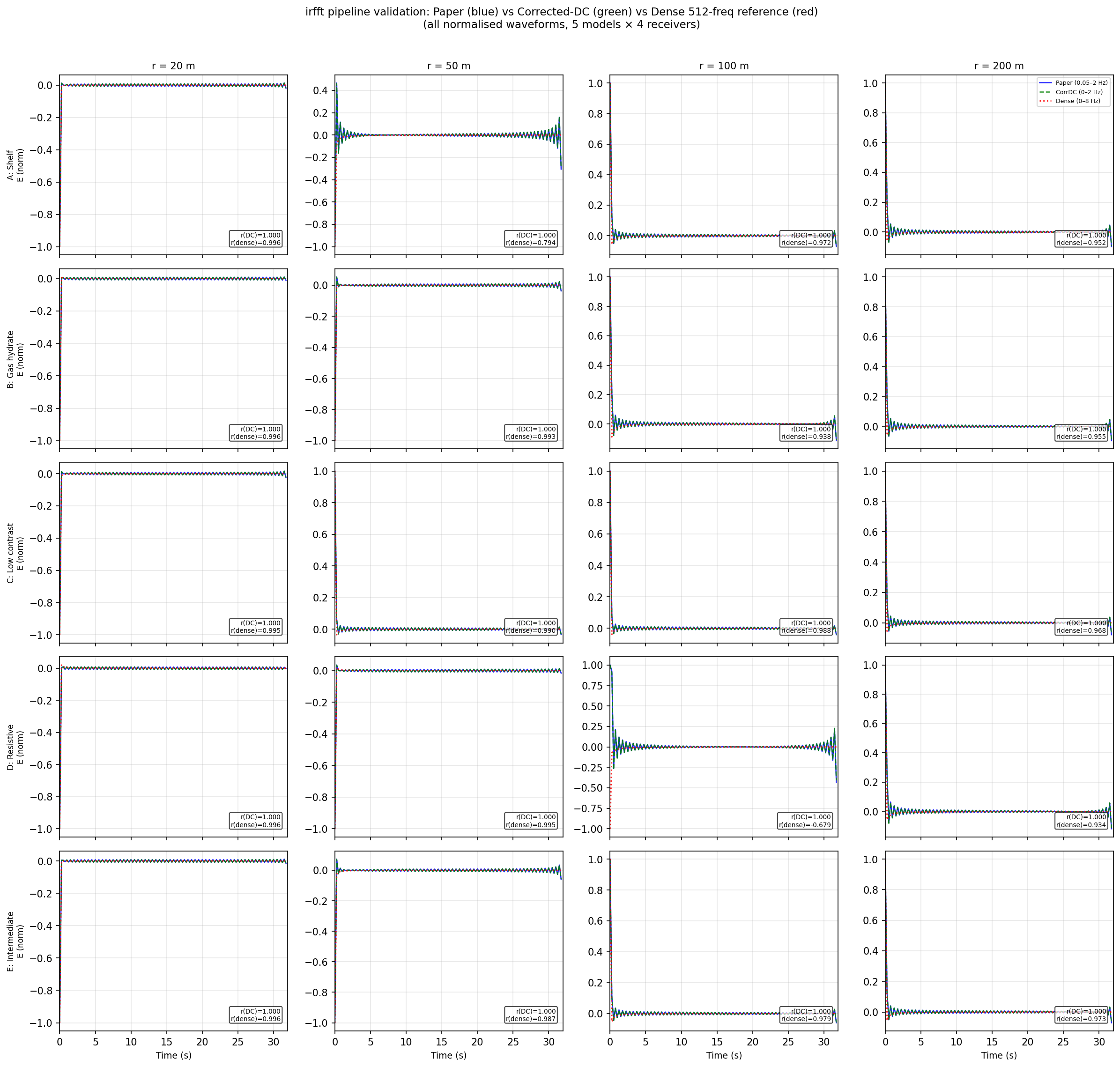}
  \caption{Validation of the irfft synthesis pipeline.  Each panel
  overlays the normalised waveform from the paper's 64-frequency
  irfft pathway (blue solid), the corrected-DC pathway (green dashed,
  frequencies 0--2\,Hz including a near-DC proxy), and a dense
  512-frequency reference (red dotted, 0--8\,Hz).  Five
  field-representative models $\times$ four receiver offsets $= 20$
  comparisons.  The paper and corrected-DC waveforms are virtually
  identical ($r \geq 0.9998$ in all cases), confirming that the DC
  misassignment is negligible.  Differences with the dense reference
  (mean $r = 0.97$ excluding one outlier) reflect bandwidth
  truncation at 2\,Hz, not a synthesis artefact.}
  \label{fig:supp_irfft}
\end{figure}

The bandwidth effect is larger but still moderate: the paper waveform
correlates at $r \geq 0.93$ with the dense reference in 18 of 20
pairs (mean RMSE $= 0.029$), with one outlier at Model~D
($\sigma_2 = 0.008$\,S\,m$^{-1}$), $r = 100$\,m, where the highly
resistive seafloor shifts significant energy above 2\,Hz.  Excluding
this outlier, the mean correlation is 0.969.  A parameter-sensitivity
test confirms that a $\pm$5\% perturbation in each of the four
earth-model parameters produces waveform changes in the same
direction in both representations, with the 0.05--2\,Hz band
retaining 25--100\% of the sensitivity of the 0--8\,Hz band for
all four parameters.  The paper's irfft pipeline thus provides a
faithful---if bandlimited---representation of the physical signal,
and the network's learned features reflect genuine earth-model
sensitivity rather than synthesis artefacts.
Fig.~\ref{fig:supp_irfft} shows the overlay for all 20 pairs.

\section{Step-Off Source Validation}
\label{sec:supp_stepoff}

DualTCN is trained on impulse-response waveforms.  Deployment on
field data acquired under the step-off convention requires a
pre-processing step: division by $\mathrm{i}\,2\pi f$ in the
frequency domain (or temporal integration) before the time series is
passed to the network.

To validate this pipeline, we synthesise step-off responses for the
five field-representative models and retransform them to the
impulse-response representation used by DualTCN.
Table~\ref{tab:supp_stepoff} compares the resulting predictions
(``RT'', retransformed from step-off) against predictions obtained
directly from impulse-response inputs (``Imp'').

\begin{table*}[t]
  \centering
  \footnotesize
  \caption{Step-off source validation: DualTCN predictions on
  impulse-response inputs (Imp) versus retransformed step-off inputs
  (RT) for five field-representative models.  ``Max \% diff'' is the
  largest absolute percentage difference between Imp and RT across all
  four parameters.  Units: $\sigma$ in S/m, $d$ in m.}
  \label{tab:supp_stepoff}
  \setlength{\tabcolsep}{3.5pt}
  \begin{tabular}{llrrrr@{\;\;}rrrr@{\;\;}r}
    \toprule
    & Model
      & $\hat\sigma_1^\mathrm{Imp}$ & $\hat\sigma_2^\mathrm{Imp}$
      & $\hat{d}_1^\mathrm{Imp}$ & $\hat{d}_2^\mathrm{Imp}$
      & $\hat\sigma_1^\mathrm{RT}$ & $\hat\sigma_2^\mathrm{RT}$
      & $\hat{d}_1^\mathrm{RT}$ & $\hat{d}_2^\mathrm{RT}$
      & Max \% \\
    \midrule
    A & Shelf
      & 3.04 & 0.440 & 80.3 & 36.2
      & 3.27 & 0.479 & 80.8 & 32.4 & 10.4 \\
    B & Gas hydrate
      & 3.19 & 0.055 & 97.1 & 18.0
      & 3.44 & 0.081 & 98.0 & 17.3 & 47.2 \\
    C & Low contrast
      & 2.51 & 0.714 & 58.5 & 43.2
      & 2.67 & 0.739 & 58.6 & 41.5 & 6.4 \\
    D & Resistive
      & 3.71 & 0.008 & 114.2 & 14.4
      & 3.98 & 0.017 & 115.5 & 12.0 & 117.3 \\
    E & Intermediate
      & 1.73 & 0.245 & 90.3 & 37.6
      & 1.84 & 0.276 & 90.9 & 36.3 & 12.6 \\
    \bottomrule
  \end{tabular}
\end{table*}

For three of the five models (A, C, E), the retransformation
introduces differences below 13\%, confirming that the pre-processing
pipeline is viable under typical shelf and sediment conditions.
Model~B (gas hydrate) shows a 47\% difference concentrated in
$\sigma_2$ (0.055 vs.\ 0.081\,S/m), where the low seafloor
conductivity amplifies relative sensitivity to the retransformation.
Model~D (resistive basement, $\sigma_2 = 0.008$\,S/m) is the worst
case: $\hat{\sigma}_2$ shifts from 0.008 to 0.017\,S/m (117\%
relative change), reflecting the fact that highly resistive targets
push significant spectral energy above the 2\,Hz bandwidth, and the
retransformation amplifies the resulting truncation artefact.

\section{Physical-Unit Error Conversion}
\label{sec:supp_phys}

Table~\ref{tab:supp_phys} translates the normalised log-space RMSE
values into physical units to aid interpretability.

\begin{table*}[t]
  \centering
  \footnotesize
  \caption{DualTCN test-set RMSE converted to physical units
  (150\,000 test samples).
  $e_\mathrm{norm}$: normalised $[0,1]$ log-space RMSE.
  $e_\mathrm{log}$: RMSE in $\log_{10}$ physical units.
  $\Gamma = 10^{e_\mathrm{log}}$: geometric RMSE factor.
  ``Physical RMSE at geo-mean'': $\Gamma - 1$ applied to the
  geometric mean of the training range.}
  \label{tab:supp_phys}
  \begin{tabular}{lcccccc}
    \toprule
    Parameter & Range & Geo-mean & $e_\mathrm{norm}$ & $e_\mathrm{log}$ & $\Gamma$ & Phys.\ RMSE \\
    \midrule
    $\sigma_1$ (S\,m$^{-1}$) & 0.1--5.0 & 0.71 & 0.020 & 0.034 & 1.08 & 0.06\,S\,m$^{-1}$ \\
    $\sigma_2$ (S\,m$^{-1}$) & 0.001--1 & 0.032 & 0.092 & 0.276 & 1.89 & 0.028\,S\,m$^{-1}$ \\
    $d_1$ (m)                & 50--150 & 87 & 0.031 & 0.015 & 1.04 & 3\,m \\
    $d_2$ (m)                & 10--50  & 22 & 0.165 & 0.115 & 1.30 & 7\,m \\
    \bottomrule
  \end{tabular}
\end{table*}

\section{Full Ablation Results}
\label{sec:supp_ablation}

Tables~\ref{tab:supp_rmse} and~\ref{tab:supp_r2} report normalised
RMSE, total loss, inference timing, and per-parameter $R^2$ for all
thirteen ablation variants.

\begin{table*}[t]
  \centering
  \footnotesize
  \caption{Normalised RMSE, total loss, and single-sample latency
  (A100 GPU, 1\,000 runs). Bold = best; $\dagger$ = worse than
  baseline.}
  \label{tab:supp_rmse}
  \begin{tabular}{llrcccccc}
    \toprule
    & Variant & Params & $\sigma_1$ & $\sigma_2$ & $d_1$ & $d_2$
              & Total & ms \\
    \midrule
    & PCRN        & 638K & 0.026 & 0.110 & 0.033 & 0.183
                  & 0.1032 & 2.6 \\
    P1  & 200 ep      & 638K & 0.027 & 0.102 & 0.033 & 0.184
        & 0.0932 & 2.5 \\
    P2  & +amp-ratio  & 638K & 0.027 & 0.101 & 0.034 & 0.192
        & 0.0959 & 2.6 \\
    P3$^\dagger$
        & $d_2$ wt=5  & 638K & 0.025 & 0.111 & 0.035 & 0.180
        & 0.1145 & 2.7 \\
    P4$^\dagger$
        & CrossRecv   & 576K & 0.036 & 0.179 & 0.036 & 0.259
        & 0.2469 & 2.1 \\
    P5a & TCN-only    & \textbf{201K} & 0.022 & 0.109
        & \textbf{0.031} & 0.187 & 0.1020 & 2.0 \\
    P5b$^\dagger$
        & TF-only     & 877K & 0.097 & 0.155 & 0.137 & 0.224
        & 0.2248 & 1.4 \\
    P5c$^\dagger$
        & MLP-only    & 952K & 0.091 & 0.269 & 0.082 & 0.272
        & 0.4450 & 0.5 \\
    P6$^\dagger$
        & Latent=512  & 770K & 0.027 & 0.105 & \textbf{0.031}
        & 0.208 & 0.1007 & 2.6 \\
    P7  & MultiScale  & 384K & 0.021 & 0.100 & \textbf{0.031}
        & 0.184 & 0.0931 & 3.8 \\
    P8  & TwoStage    & 401K & 0.020 & 0.098 & 0.033 & 0.185
        & 0.0892 & 3.8 \\
    P10a$^\dagger$
        & iTransf.    & 381K & 0.040 & 0.180 & 0.041 & 0.220
        & 0.2204 & 1.0 \\
    P10b$^\dagger$
        & PatchTST    & 369K & 0.050 & 0.202 & 0.050 & 0.248
        & 0.2749 & 1.1 \\
    \textbf{DualTCN}
        &             & 379K & \textbf{0.020} & \textbf{0.092}
        & \textbf{0.031} & \textbf{0.165}
        & \textbf{0.0771} & 3.5 \\
    \bottomrule
  \end{tabular}
\end{table*}

\begin{table*}[t]
  \centering
  \footnotesize
  \caption{Per-parameter $R^2$ and batch-256 throughput.
  Bold = best.}
  \label{tab:supp_r2}
  \begin{tabular}{llccccr}
    \toprule
    & Variant & $R^2_{\sigma_1}$ & $R^2_{\sigma_2}$ & $R^2_{d_1}$
              & $R^2_{d_2}$ & Throughput \\
    \midrule
    & PCRN        & 0.992 & 0.855 & 0.986 & 0.541 &  45\,K/s \\
    P1  & 200 ep      & 0.991 & 0.876 & 0.986 & 0.536
        &  44\,K/s \\
    P2  & +amp-ratio  & 0.991 & 0.877 & 0.985 & 0.496
        &  45\,K/s \\
    P3  & $d_2$ wt=5  & 0.992 & 0.853 & 0.984 & 0.560
        &  44\,K/s \\
    P4  & CrossRecv   & 0.984 & 0.615 & 0.984 & 0.082
        &  59\,K/s \\
    P5a & TCN-only    & 0.994 & 0.857 & 0.987 & 0.525
        & 127\,K/s \\
    P5b & TF-only     & 0.886 & 0.713 & 0.760 & 0.318
        &  34\,K/s \\
    P5c & MLP-only    & 0.901 & 0.133 & 0.915 & $-$0.008
        & 550\,K/s \\
    P6  & Latent=512  & 0.991 & 0.868 & 0.987 & 0.411
        &  45\,K/s \\
    P7  & MultiScale  & 0.995 & 0.879 & \textbf{0.988} & 0.539
        &  67\,K/s \\
    P8  & TwoStage    & 0.995 & 0.885 & 0.986 & 0.532
        &  67\,K/s \\
    P10a & iTransf.   & 0.980 & 0.610 & 0.979 & 0.342
        & \textbf{247\,K/s} \\
    P10b & PatchTST   & 0.970 & 0.512 & 0.968 & 0.162
        & 69\,K/s \\
    \textbf{DualTCN}
        &             & \textbf{0.995} & \textbf{0.898} & 0.987
        & \textbf{0.627} & 76\,K/s \\
    \bottomrule
  \end{tabular}
\end{table*}

\section{Why Attention-Based Encoders Underperform}
\label{sec:supp_transformer}

Both P10a (iTransformer) and P10b (PatchTST) produce total losses
above 0.22---more than $2.8\times$ worse than DualTCN and worse than
the plain TCN-only baseline.

\subsection{Specific Configurations Tested}

We ran exploratory variants of the Transformer-only encoder (P5b)
with 1, 2, and 4 layers and 2, 4, and 8 heads (all within
$\pm 0.01$ total loss of the reported P5b).  Beyond two layers,
training instability (gradient-norm spikes) was observed.
Sinusoidal, learnable absolute, and relative (RoPE-style) positional
encodings were compared; none materially affected the final loss.
For PatchTST, patch lengths $\{8, 16, 32\}$ were tested; smaller
patches marginally improved $R^2_{\sigma_1}$ but not $R^2_{d_2}$.

\subsection{Physical Interpretation}

The core difficulty is inductive bias, not hyperparameters.  The
MCSEM transient is not exchangeable in time: the first 32 samples
encode seawater properties through a distinct diffusion regime,
while the last 64 encode the seafloor through exponential decay.
Global self-attention treats all positions as equally relevant,
which conflates these two physically separate regimes.

The iTransformer collapses each channel to a single token, discarding
temporal ordering entirely ($R^2_{d_2} = 0.342$).  PatchTST retains
locality within each patch but the channel-independent design delays
cross-receiver integration.  By contrast, DualTCN hardcodes the known
temporal structure with a separate encoder branch for the late-time
window.

\section{Full Uncertainty Quantification Comparison}
\label{sec:supp_uq}

Four UQ methods are compared on 5\,000 held-out test samples:
MC-Dropout ($T = 100$ passes, $p = 0.30$), temperature-scaled
MC-Dropout, split conformal prediction ($n_\mathrm{cal} = 75{,}000$),
and a five-member deep ensemble.

MC-Dropout keeps dropout active at inference and constructs Gaussian
intervals from the predictive mean and standard deviation.
Temperature scaling learns one scalar $\tau_k$ per parameter on the
validation set to dilate the MC-Dropout variance.  Split conformal
prediction uses the empirical residual quantile to construct intervals
with provable marginal coverage.  The deep ensemble aggregates five
independently trained DualTCN models (seeds 0--4).

Table~\ref{tab:supp_uq} reports the full PICP at six nominal levels,
MPIW$_{90}$, and point-estimate RMSE.

\begin{table}[t]
  \centering
  \footnotesize
  \caption{PICP at nominal levels $\alpha \in \{50, 70, 80, 90, 95,
  99\%\}$ for four UQ methods (5\,000 held-out test samples).
  Bold = closest to nominal for $\sigma_2$ and $d_2$.}
  \label{tab:supp_uq}
  \scriptsize
  \setlength{\tabcolsep}{2pt}
  \begin{tabular}{llcccc}
    \toprule
    $\alpha$ & Method & $\sigma_1$ & $\sigma_2$ & $d_1$ & $d_2$ \\
    \midrule
    50\% & MC-Dropout      & 0.688 & 0.434 & 0.291 & 0.280 \\
         & Temp-scaled     & 0.552 & 0.609 & 0.488 & 0.585 \\
         & Split conformal & 0.504 & \textbf{0.480} & 0.489 & \textbf{0.496} \\
         & Deep ensemble   & 0.267 & 0.214 & 0.149 & 0.140 \\
    \midrule
    90\% & MC-Dropout      & \textbf{0.944} & 0.664 & 0.596 & 0.572 \\
         & Temp-scaled     & 0.863 & 0.845 & 0.905 & 0.854 \\
         & Split conformal & 0.906 & \textbf{0.900} & \textbf{0.902} & \textbf{0.905} \\
         & Deep ensemble   & 0.581 & 0.480 & 0.278 & 0.362 \\
    \midrule
    99\% & MC-Dropout      & 0.990 & 0.785 & 0.827 & 0.715 \\
         & Temp-scaled     & 0.956 & 0.940 & 0.987 & 0.947 \\
         & Split conformal & \textbf{0.993} & \textbf{0.989} & \textbf{0.990} & \textbf{0.991} \\
         & Deep ensemble   & 0.764 & 0.656 & 0.366 & 0.559 \\
    \midrule
    \multicolumn{6}{l}{\textit{MPIW$_{90}$ (normalised)}} \\
    & MC-Dropout      & 0.078 & 0.077 & 0.059 & 0.135 \\
    & Temp-scaled     & 0.055 & 0.149 & 0.112 & 0.342 \\
    & Split conformal & 0.062 & 0.306 & 0.096 & 0.500 \\
    & Deep ensemble   & 0.026 & 0.074 & 0.018 & 0.125 \\
    \bottomrule
  \end{tabular}
\end{table}

\section{Ablation Variant Descriptions}
\label{sec:supp_variants}

\begin{table}[t]
  \centering
  \footnotesize
  \caption{Thirteen ablation variants and their modifications
  relative to the PCRN baseline.}
  \label{tab:supp_experiments}
  \setlength{\tabcolsep}{2pt}
  \begin{tabular}{lp{4.5cm}}
    \toprule
    ID & Modification \\
    \midrule
    P1   & Doubled training (200 epochs) \\
    P2   & Three inter-receiver log-amplitude ratios (11 ch) \\
    P3   & $d_2$ loss weight $\times 5$ \\
    P4   & Per-receiver TCN + cross-attention fusion \\
    P5a  & TCN-only encoder \\
    P5b  & Transformer-only encoder \\
    P5c  & Flat MLP encoder \\
    P6   & Latent dimension doubled to 512 \\
    P7   & Multi-scale TCN (3 dilation schedules) \\
    P8   & Two-stage cascaded prediction \\
    P10a & iTransformer \\
    P10b & PatchTST \\
    DualTCN & Late-time branch + $d_\mathrm{sf}$ head \\
    \bottomrule
  \end{tabular}
\end{table}

\section{Amplitude-Channel Perturbation Results}
\label{sec:supp_amp}

\begin{table}[t]
  \centering
  \footnotesize
  \caption{DualTCN $R^2$ under amplitude-channel perturbations
  (150\,000 test samples).  Panel~(A): random noise.
  Panel~(B): systematic bias.  Bold = clean baseline.}
  \label{tab:supp_amp}
  \setlength{\tabcolsep}{2pt}
  \begin{tabular}{l@{\;\;}ccccc}
    \toprule
    \multicolumn{6}{c}{\textbf{(A) Random noise}} \\
    $\sigma_\mathrm{amp}$ &
      $R^2_{\sigma_1}$ & $R^2_{\sigma_2}$ & $R^2_{d_1}$ & $R^2_{d_2}$ & $\bar{R}^2$ \\
    \midrule
    \textbf{0.00} & \textbf{0.995} & \textbf{0.898} & \textbf{0.987} & \textbf{0.627} & \textbf{0.877} \\
    0.01 & 0.919 & 0.380 & 0.939 & $-$0.784 & 0.363 \\
    0.02 & 0.784 & $-$0.112 & 0.803 & $-$1.465 & 0.002 \\
    0.05 & 0.368 & $-$1.149 & 0.280 & $-$2.124 & $-$0.656 \\
    0.10 & $-$0.130 & $-$1.941 & $-$0.558 & $-$2.394 & $-$1.256 \\
    0.20 & $-$0.683 & $-$2.431 & $-$1.510 & $-$2.671 & $-$1.824 \\
    \midrule
    \multicolumn{6}{c}{\textbf{(B) Systematic bias}} \\
    $\beta$ &
      $R^2_{\sigma_1}$ & $R^2_{\sigma_2}$ & $R^2_{d_1}$ & $R^2_{d_2}$ & $\bar{R}^2$ \\
    \midrule
    $-$0.20 & 0.746 & $-$0.121 & 0.980 & 0.494 & 0.525 \\
    $-$0.10 & 0.958 & 0.368 & 0.990 & 0.601 & 0.729 \\
    \textbf{0.00} & \textbf{0.995} & \textbf{0.898} & \textbf{0.987} & \textbf{0.627} & \textbf{0.877} \\
    $+$0.10 & 0.951 & 0.626 & 0.985 & 0.273 & 0.709 \\
    $+$0.20 & 0.822 & 0.351 & 0.974 & $-$0.079 & 0.517 \\
    \bottomrule
  \end{tabular}
\end{table}

\section{Full Conventional Inversion Benchmark}
\label{sec:supp_benchmark}

\begin{table*}[t]
  \centering
  \caption{DualTCN versus conventional inversion
  (same \texttt{empymod} forward operator).  All multi-start
  runs use 8 starts.  Bold = best in each column.}
  \label{tab:supp_benchmark}
  \footnotesize
  \begin{tabular}{llccccccc}
    \toprule
    & Method & $R^2_{\sigma_1}$ & $R^2_{\sigma_2}$ & $R^2_{d_1}$
             & $R^2_{d_2}$ & $\bar{R}^2$
             & s/sample & Evals \\
    \midrule
    \multicolumn{9}{l}{\textit{Untuned baselines (500 samples)}} \\
    & NLS-LM (1 start)       & 0.592 & $-$0.292 & 0.378 & $-$0.553 & 0.031 & 2.75 & 49 \\
    & NLS-LBFGSB (1 start)   & 0.465 & $-$1.077 & 0.337 & $-$0.927 & $-$0.301 & 8.01 & 291 \\
    & NLS-LM (8 starts)      & 0.938 & $-$0.330 & 0.961 & $-$1.053 & 0.129 & 12.09 & 338 \\
    & NLS-LBFGSB (8 starts)  & 0.961 & 0.179 & 0.980 & $-$0.362 & 0.439 & 73.58 & 2{,}630 \\
    \midrule
    \multicolumn{9}{l}{\textit{Occam regularised (200 samples)}} \\
    & Occam-LM ($\lambda$=0.01) & 0.730 & $-$0.385 & 0.853 & $-$1.030 & 0.042 & 14.71 & 170 \\
    \midrule
    \multicolumn{9}{l}{\textit{Improved baselines (500 samples)}} \\
    & TT-LM                  & 0.938 & $-$0.330 & 0.961 & $-$1.053 & 0.129 & 13.00 & 378 \\
    & TIK-LBFGSB             & 0.962 & 0.263 & 0.982 & $-$0.296 & 0.478 & 123.14 & 4{,}417 \\
    & WS-LM (DualTCN init)   & 0.993 & 0.689 & 0.997 & 0.343 & 0.756 & 13.22 & 380 \\
    & WS-LBFGSB (DualTCN init) & 0.993 & 0.781 & \textbf{0.998} & 0.364 & 0.784 & 92.80 & 3{,}312 \\
    \midrule
    \multicolumn{9}{l}{\textit{Neural network}} \\
    & \textbf{DualTCN}       & \textbf{0.996} & \textbf{0.884} & 0.988 & \textbf{0.578} & \textbf{0.862} & \textbf{0.0035} & 1 \\
    \bottomrule
  \end{tabular}
\end{table*}

\section{Complete Training Specification}
\label{sec:supp_training}

This section provides the full specification of the data generation,
input representation, and training protocol to facilitate
reproducibility.

\subsection*{S10.1~~Forward Model and Geometry}

\begin{itemize}[leftmargin=1.5em, itemsep=2pt]
  \item \textbf{Forward operator}: \texttt{empymod} v2.3
    \citep{werthmullerEmpymod}, 1D semi-analytic (digital-filter
    Hankel transform).  Strictly 1D layered earth; no 2D/3D effects.
  \item \textbf{Earth model}: 3 layers --- air ($\rho \to \infty$),
    seawater ($\sigma_1$), uniform seafloor half-space ($\sigma_2$).
  \item \textbf{Source}: unit-moment horizontal electric dipole (HED)
    at position $(0, 0, d_1)$ metres below the sea surface, oriented
    in the $x$-direction (inline).
  \item \textbf{Receivers}: 4 inline receivers at horizontal offsets
    $r \in \{20, 50, 100, 200\}$\,m from the source, all at depth
    $z_\mathrm{obs} = 20$\,m below the sea surface.  The receiver
    component is $E_x$ (inline electric field).
  \item \textbf{Frequency sampling}: 64 frequencies linearly spaced
    from $f_\mathrm{min} = 0.05$\,Hz to $f_\mathrm{max} = 2.0$\,Hz
    ($\Delta f \approx 0.031$\,Hz).  Linear (not logarithmic)
    spacing.
  \item \textbf{Time-domain conversion}: \texttt{numpy.fft.irfft}
    with output length $N = 128$, yielding $\Delta t = 0.25$\,s and
    a 32\,s time window.  No windowing or tapering applied; the
    0.05\,Hz response occupies the DC bin (see main text
    Section~3.1 for discussion).
\end{itemize}

\subsection*{S10.2~~Parameter Sampling}

All four parameters are sampled \emph{independently} (no joint
correlations) using the following distributions:

\begin{itemize}[leftmargin=1.5em, itemsep=2pt]
  \item $\sigma_1$: log-uniform on $[10^{-1.0}, 10^{0.7}]$ =
    $[0.10, 5.01]$\,S/m.
  \item $\sigma_2$: log-uniform on $[10^{-3.0}, 10^{0.0}]$ =
    $[0.001, 1.0]$\,S/m.
  \item $d_1$: uniform on $[50, 150]$\,m (physical space, not
    log-space).
  \item $d_2$: uniform on $[10, 50]$\,m (physical space).
  \item $v_0$ (source velocity, Doppler correction): uniform on
    $[0, 100]$\,m/s.
\end{itemize}

One million samples are drawn with seed 42
(\texttt{numpy.random.default\_rng(42)}).  The five
field-representative models (Section~4.6 of the main text) fall
within the training bounds for $\sigma_1$, $d_1$, and $d_2$; Models
B and D have $\sigma_2 = 0.05$ and $0.008$\,S/m, respectively---both
within the training range but in the sparsely sampled low-$\sigma_2$
tail.

\subsection*{S10.3~~Input Representation and Normalisation}

For each receiver $j \in \{1, \ldots, 4\}$:
\begin{enumerate}[leftmargin=1.5em, itemsep=2pt]
  \item Compute the complex frequency-domain response
    $E_j(f_k)$ via empymod for $k = 1, \ldots, 64$.
  \item Apply irfft to obtain $g_j(t_n)$ for $n = 0, \ldots, 127$.
  \item Record the peak amplitude:
    $A_j = \max_n |g_j(t_n)|$.
  \item Peak-normalise the waveform:
    $\tilde{g}_j(t_n) = g_j(t_n) / A_j$.
  \item Compute $\ell_j = \log_{10}(A_j)$.
  \item Construct two input channels:
    channel $2j{-}1 = \tilde{g}_j(t_n)$ (128 samples),
    channel $2j = \ell_j$ broadcast to 128 samples.
\end{enumerate}
The resulting input tensor has shape $(8, 128)$.
Target parameters are normalised to $[0, 1]$ in $\log_{10}$ space
using the bounds in Table~1 of the main text.

\subsection*{S10.4~~Noise Injection Protocol}

During training only (validation and test sets are clean unless
otherwise noted):
\begin{itemize}[leftmargin=1.5em, itemsep=2pt]
  \item \textbf{Waveform noise (all variants)}: additive Gaussian
    noise $\epsilon_n \sim \mathcal{N}(0, \sigma_w^2)$ added to
    the normalised waveform channels only.  The standard deviation
    $\sigma_w$ is drawn log-uniformly from $[10^{-3}, 10^{-1}]$ per
    sample (corresponding to $\sim$11--51\,dB SNR).  The
    log-amplitude channels are \emph{not} perturbed.
  \item \textbf{Amplitude noise (DualTCN-AmpAug only)}: curriculum
    training.  Epochs 1--20: clean.  Epochs 20--40: linear ramp of
    $\sigma_\mathrm{amp}$ from 0 to the target range.  Epochs
    40--100: $\sigma_\mathrm{amp} \sim U[0.001, 0.01]$ in
    $\log_{10}$ units applied independently per receiver.
  \item \textbf{Colored noise (DualTCN-Colored only)}: $1/f$
    (pink) noise spectrum replacing white Gaussian on the waveform
    channels, generated via spectral shaping.
  \item \textbf{Per-receiver bias (DualTCN-RecvBias only)}: additive
    bias $\beta_j \sim U[-0.03, +0.03]$ $\log_{10}$ applied
    independently per receiver to the log-amplitude channels.
\end{itemize}

\subsection*{S10.5~~Training Hyperparameters}

\begin{itemize}[leftmargin=1.5em, itemsep=2pt]
  \item \textbf{Optimiser}: AdamW ($\beta_1 = 0.9$, $\beta_2 = 0.999$,
    $\lambda = 10^{-3}$, weight decay $10^{-3}$).
  \item \textbf{Learning rate schedule}: one-cycle cosine
    (5-epoch linear warmup, peak rate $5 \times 10^{-4}$, final rate
    $5 \times 10^{-6}$; final-div-factor = 100).
  \item \textbf{Batch size}: 256.
  \item \textbf{Epochs}: 100 (no explicit early stopping; best
    validation-loss checkpoint selected).
  \item \textbf{Dataset split}: 700\,K train / 150\,K val / 150\,K
    test (70/15/15, sequential split).
  \item \textbf{Loss}: $\mathcal{L}_\mathrm{MSE}^\mathrm{prof}
    + 2 \sum_{i=1}^{4} w_i \mathcal{L}_\mathrm{Huber}^{(i)}$;
    Huber $\delta = 0.1$; weights
    $\mathbf{w} = [1, 3, 3, 2]$ for
    $[\sigma_1, \sigma_2, d_1, d_2]$.
  \item \textbf{Gradient clipping}: max norm = 1.0.
  \item \textbf{Hardware}: single NVIDIA A100-SXM4-40\,GB GPU.
  \item \textbf{Training time}: $\sim$3--4 GPU-hours per variant.
  \item \textbf{Framework}: PyTorch 2.x; \texttt{torch.compile} and
    automatic mixed precision (AMP) enabled.
  \item \textbf{DualTCN auxiliary loss}: Huber($\hat{d}_\mathrm{sf}$,
    $d_\mathrm{sf}^\mathrm{true}$) with weight 0.5, added to the
    main loss.  $d_\mathrm{sf}$ normalised as
    $(\log_{10}(d_\mathrm{sf}) - 1.778) / 0.523$.
\end{itemize}

\subsection*{S10.6~~Domain of Applicability}

DualTCN's predictions are valid under the following conditions:
\begin{enumerate}[leftmargin=1.5em, itemsep=2pt]
  \item The subsurface is well approximated by a laterally uniform
    1D layered model beneath the receiver spread ($< 200$\,m lateral
    extent).  Lateral heterogeneity not represented in training.
  \item Earth-model parameters fall within the training ranges
    (Table~1).  Extrapolation beyond these ranges is untested.
  \item The acquisition geometry matches the training configuration
    (four inline receivers at 20--200\,m offsets, 20\,m depth).
    Different geometries require retraining on matched synthetics.
  \item The source waveform is either impulse (direct compatibility)
    or step-off (requires $1/(\mathrm{i}2\pi f)$ pre-processing;
    validated for $\sigma_2 \geq 0.05$\,S/m, degraded for
    resistive targets).
  \item Amplitude calibration uncertainty is $\leq$2\% per receiver
    (for the unaugmented variant) or $\leq$5\% (for DualTCN-AmpAug).
\end{enumerate}

\section{Extended Background and Related Work}
\label{sec:supp_background}

This section provides the full background and related work discussion
that was condensed in the main text.

\subsection*{S11.1~~Sensitivity Structure of MCSEM Data}

How much information an MCSEM measurement carries about the
subsurface depends on both the source--receiver offset and the time
window being examined. Receivers close to the source are dominated by
the direct arrival and the airwave, so they tell us relatively little
about what lies beneath the seafloor. Receivers at larger offsets, by
contrast, record the guided wave that has interacted with the
resistive sub-seafloor, and they can therefore resolve conductivity
contrasts at depth \citep{key2012marine}.

In the time domain, this offset dependence translates into a temporal
hierarchy. Early-time samples are sensitive mainly to the water column
(its conductivity $\sigma_1$ and depth $d_1$), while late-time
samples---where the signal decays approximately as
$\exp(-2 u_1 d_2)$, with $u_1$ the vertical wavenumber in
seawater---carry information about the seafloor ($\sigma_2$ and the
source-to-seafloor distance $d_2$). Recording the full transient preserves
information across a continuous range of diffusion times, in contrast
to frequency-domain measurements, which sample only a discrete set
of frequencies \citep{constable2010ten}.

This temporal hierarchy is central to the architectural decisions we
make in this work. If different time windows contain information about
different parameters, then a network that treats all time samples
identically is leaving information on the table; a better design
would give the network the ability to specialise.

\subsection*{S11.2~~Classical 1D MCSEM Inversion and Acquisition Variants}

The gold standard for 1D MCSEM inversion is the Occam algorithm
\citep{constable1987occam}, which minimises a data misfit subject to
a smoothness constraint on the resistivity--depth profile, producing
the simplest model consistent with the data.  One-dimensional Occam
inversion of multi-component, multi-frequency CSEM data was
formalised by \citet{key2009}, who provided analytic expressions for
the Jacobians of the 1D layered forward operator that make the
inversion computationally efficient and accurate.  The two-layer
parameterisation used in this study (seawater conductivity $\sigma_1$,
seafloor conductivity $\sigma_2$, source depth $d_1$, gap
$d_2$) is precisely the minimal model that Occam reduces to when
applied to shallow-water, single-interface geology; it is also the
standard starting model for more complex iterative schemes
\citep{constable2006, key2012marine}.

A practical limitation of Occam and related regularised methods is
computational cost: even with analytic Jacobians each sample requires
dozens to hundreds of forward evaluations, making real-time
interpretation during acquisition infeasible.
\citet{constable2006} and \citet{key2012marine} discuss these
operational constraints and the consequent demand for faster
inversion strategies.

The majority of published MCSEM surveys use a
\emph{seabed-deployed} geometry: receivers are placed on the ocean
floor and the source is towed just above them.  A geometrically
distinct variant, \emph{towed-streamer EM} (TSEM), tows both source
and receivers at the sea surface \citep{ziolkowski2007}, trading the
deep-seafloor coupling of the seabed geometry for continuous coverage
and rapid turnaround.

DualTCN is trained and evaluated using a \emph{shallow-tow}
configuration in which the receivers are modelled at a fixed depth of
$z_\mathrm{obs} = 20$\,m below the sea surface---i.e., within the
water column, not on the seafloor.  This is an intermediate geometry
between seabed-deployed and surface-towed: the source operates at
50--150\,m depth (within or near the seafloor interface) while the
receivers are above it, at horizontal offsets of 20--200\,m.
The electromagnetic field of a horizontal electric dipole (HED)
moving at constant speed within such a layered conducting medium---a
horizontal layer bounded by two homogeneous half-spaces representing
seawater and seafloor---was analysed by \citet{sami2015motion}, whose
formulation provides the physical basis for the source model used
here.
Compared with seabed receivers ($z_\mathrm{obs} \approx
d_\mathrm{sf}$), water-column receivers at 20\,m depth are further
from the seafloor interface and closer to the air--sea boundary,
which reduces the amplitude of the seafloor-refracted signal and
increases sensitivity to the airwave, particularly at near offsets.
This geometry choice was motivated by the computational simplicity
of a single, fixed observation depth, but the sensitivity differences
relative to a seabed-deployed survey should be borne in mind when
interpreting the reported accuracies.  The neural network framework
itself generalises to any source--receiver geometry via retraining on
geometry-matched synthetic data; applying DualTCN to a seabed-deployed
or TSEM configuration would require regenerating the training set with
the appropriate $z_\mathrm{obs}$ and offset ranges.

\subsection*{S11.3~~Deep Learning for Electromagnetic Inversion}

Deep learning has been applied to electromagnetic inversion with
increasing scope.  \citet{puzyrev2019deep} demonstrated CNN-based
1D frequency-domain CSEM inversion; \citet{puzyrev2021multi}
extended this to joint frequency- and time-domain frameworks.
\citet{araya2018deep} showed that neural-network initial models
reduce conventional solver iterations from fifty to single digits,
and \citet{liu2020deep} achieved competitive accuracy for resistivity
inversion at a fraction of the cost.  \citet{zhang2024_3d} trained
3D convolutional models on $> 500{,}000$ CSEM samples.

In airborne time-domain EM (ATEM), physics-guided strategies closely
related to DualTCN have emerged: differentiable physics decoders,
disentangled multi-branch encoders, and auxiliary depth losses
\citep{moghadas2020deep, liu2021atem, colombo2021deep}.
DualTCN adapts these ideas to the marine CSEM setting
(Section~S16).  Dilated TCNs
\citep{bai2018empirical} provide the temporal inductive bias, and
physics-informed neural networks \citep{raissi2019physics} offer a
complementary approach.

Beyond supervised networks, several alternative paradigms have
emerged.  Projected Gauss--Newton methods with learned priors
\citep{abubakar2012pgn} integrate data-driven regularisation into
iterative solvers, retaining explicit data-misfit control at the cost
of per-sample iteration.  Test-time implicit-prior methods such as
Deep Image Prior (DIP) \citep{ulyanov2018dip} have been adapted for
geophysical inversion, producing regularised estimates without labelled
training data; \citet{sun2023dip_mt} demonstrated this for
magnetotelluric (MT) inversion, and unsupervised differentiable
forward-operator approaches have been explored for MT by
\citet{bai2024diffmt}.  Ensemble Kalman inversion
\citep{iglesias2013eki} provides a derivative-free alternative
for low-dimensional inverse problems, naturally producing
uncertainty estimates through ensemble spread; it is particularly
well suited to compact 1D parameterisations similar to ours but
scales poorly to the million-sample throughput required for real-time
survey processing.
The OpenEM initiative \citep{openem2021} provides public EM
datasets and benchmark models, though its current scope emphasises
onshore and airborne regimes rather than the marine time-domain
setting targeted here.
For uncertainty quantification, invertible
neural networks \citep{ardizzone2019inn}, normalising flows
\citep{papamakarios2021normalizing, bloem2023posterior}, and deep
ensembles \citep{lakshminarayanan2017} represent the current state
of the art for posterior estimation in geophysical inversion.

\subsection*{S11.4~~Open Questions}

Six significant gaps remain in the literature despite the progress
described above.

The first is that every deep-learning marine CSEM inversion published
to date works in the frequency domain
\citep{puzyrev2019deep, puzyrev2021multi, zhang2024_3d,
zhanghu2024physics, li2025marine}.
Time-domain MCSEM data, which
record causal transient responses and are increasingly used in
shallow-water and pulsed-source configurations, have not been
addressed.  (\citet{puzyrev2021multi} includes a time-domain
component, but it targets land-based TEM, not marine CSEM.)
The causal, multi-channel structure of these transients is
well suited to temporal convolutional architectures, but this
connection has not been exploited.

The second gap concerns the output representation. Existing methods
uniformly predict a discretised resistivity profile consisting of 50
to 200 independent depth samples, an output space that admits
non-monotonic and geologically implausible profiles. For the
two-layer earth model we consider here, just four scalars fully
specify the subsurface. No prior work has regressed these parameters
directly or enforced the resulting profile structure through a
differentiable analytic decoder.

Third, no systematic comparison of encoder architectures---dilated
TCN, global Transformer, flat multilayer perceptron (MLP)---has been carried out for
multi-receiver time-domain MCSEM inversion. Most studies adopt a
single architecture without empirical justification.

Fourth, the source-to-seafloor distance $d_2$ is encoded mainly in
the late-time diffusion tail. No existing architecture dedicates a
separate encoder branch to this regime, and no method has introduced
auxiliary physical objectives (such as the total seafloor depth
$d_\mathrm{sf} = d_1 + d_2$) to improve the gradient signal for
$d_2$.

Fifth, although physics-guided deep-learning strategies developed
in the ATEM community (disentangled multi-window encoders, auxiliary
depth losses, differentiable physics decoders) are directly relevant
to marine MCSEM, no study has transferred or evaluated these ideas in
the marine setting.

Sixth, all existing deep-learning MCSEM inversion methods produce
point estimates without calibrated uncertainty.  Probabilistic methods
capable of characterising the full posterior---invertible neural
networks, conditional normalising flows, and deep ensembles---have
been demonstrated for seismic and potential-field inversion
but have not been applied to marine MCSEM.

DualTCN addresses gaps one through four directly, draws on ATEM
methodology to address gap five, and provides a first MC-Dropout
uncertainty baseline---together with post-hoc calibration and
conformal prediction comparisons---that motivates the probabilistic
extensions identified in gap six.

\section{Extended Earth Model and Forward Modeling Details}
\label{sec:supp_earthmodel}

This section provides the detailed forward modeling content that was
condensed in the main text (Section~3.1).

\subsection*{S12.1~~Frequency Band}

The 0.05--2.0\,Hz band is chosen to span the diffusion depth range
relevant to our geometry and parameter space.  The electromagnetic
skin depth in a medium of conductivity $\sigma$ at frequency $f$ is
$\delta = \sqrt{2/(\omega\mu_0\sigma)}$.  For seawater
($\sigma_1 \approx 3$\,S\,m$^{-1}$) at $f_\mathrm{min} = 0.05$\,Hz,
$\delta \approx 1{,}300$\,m---well beyond the seafloor depths
considered ($d_\mathrm{sf} \leq 200$\,m), ensuring the lowest
frequency captures the full late-time diffusion signature
that carries information about $\sigma_2$ and $d_2$.  At
$f_\mathrm{max} = 2.0$\,Hz in seawater, $\delta \approx 205$\,m,
adequate to resolve the early-time interactions with the shallowest
sources considered ($d_1 \geq 50$\,m).  Frequencies above 2\,Hz
are dominated by the direct arrival and the geometric-spreading
regime rather than the diffusion tail, and carry diminishing
information about the subsurface at offsets of 20--200\,m; the
band is therefore both physically motivated and computationally
efficient.

\subsection*{S12.2~~DC Component and High-Frequency Truncation}

The lowest computed frequency is $f_\mathrm{min} = 0.05$\,Hz,
not DC ($f = 0$).  When passed to \texttt{irfft}, the bin at index
$k = 0$ is interpreted as the DC response; it instead holds the
$0.05$\,Hz response.  The true galvanic (DC) field, which arises
from charge accumulation at conductivity contrasts, is therefore
absent from the synthesised time series.  At the source--receiver
offsets used here (20--200\,m) and source depths
($d_1 = 50$--$150$\,m), the DC field is small relative to the
inductive component across our conductivity range, so this
approximation has a minor effect on the field amplitude at
$t \ll 1/f_\mathrm{min} = 20$\,s.  The spectral truncation at
2\,Hz acts as an ideal low-pass filter, suppressing content
above the Nyquist frequency.  The corresponding rectangular
spectral window creates Gibbs-like ringing at early times
($t \lesssim \Delta t = 0.25$\,s); because both the training
signals and any evaluation signal are constructed identically, the
network adapts to this representation and the ringing is not
a source of bias.

\subsection*{S12.3~~Wrap-Around and Aliasing}

The \texttt{irfft} output is a discrete, periodic sequence with
period $T = N\,\Delta t = 32$\,s.  If the transient has not
decayed to negligible amplitude by $t = 32$\,s, the periodic
extension causes temporal aliasing.  The late-time MCSEM transient
decays as $E(t) \sim t^{-5/2}$ for a dipole source in a uniform
half-space; at the receiver offsets considered (20--200\,m), the
signal amplitude at $t = 32$\,s is of order
$(32\,\mathrm{s}/t_\mathrm{peak})^{-5/2}$ times the peak value,
where $t_\mathrm{peak}$ is at most a few seconds.  This gives a
fractional wrap-around amplitude below $10^{-3}$ for all parameter
combinations in our training set, confirming that aliasing is
negligible.

\subsection*{S12.4~~Implicit Source Waveform}

\texttt{empymod} computes $E(\mathbf{r}, f)$, the complex
frequency-domain electric field for a unit-moment harmonic dipole
source.  Applying \texttt{irfft} to this spectrum (without
dividing by $\mathrm{i}\,2\pi f$) produces the time-domain
impulse response $g(t)$---the field generated by a broadband
Dirac-delta current pulse.  This differs from the step-off
convention widely used in time-domain MCSEM practice
\citep{constable2010ten}, for which the transient is
$E_\mathrm{step}(t) = \int_t^\infty g(\tau)\,\mathrm{d}\tau
= \mathcal{F}^{-1}\!\bigl[E(f)/(i2\pi f)\bigr]$.
The impulse-response convention is self-consistent across training
and evaluation---every sample is synthesised and inverted using
the same procedure---so the network correctly learns to invert
this representation.  Deployment on field data acquired under a
step-off or square-wave source protocol would require a
pre-processing step (temporal integration of the received
transient, or equivalently division by $i2\pi f$ in the frequency
domain) before the time series is passed to DualTCN;
Section~S2 validates this pipeline on five
field-representative models.

\subsection*{S12.5~~Synthesis Pipeline: Known Simplifications}

Three deliberate simplifications distinguish the synthesis pipeline
from a production MCSEM processing flow and should be borne in mind
when interpreting the reported accuracies.
\emph{First}, the DC component is absent: the lowest computed
frequency is $f_\mathrm{min} = 0.05$\,Hz, so the galvanic response
is not modelled.  The supplementary validation (Section~S1) confirms
the effect is negligible ($r \geq 0.9998$ versus a corrected-DC
reference).
\emph{Second}, the 2\,Hz bandwidth truncation acts as a low-pass
filter; a dense 8\,Hz reference shows moderate differences (mean
$r = 0.97$), with the largest impact on highly resistive targets
($\sigma_2 < 0.01$\,S\,m$^{-1}$) where significant energy lies
above 2\,Hz.  A production system targeting resistive reservoirs
should extend the band to 4--8\,Hz.
\emph{Third}, the receiver geometry ($z_\mathrm{obs} = 20$\,m,
offsets 20--200\,m) differs from standard seabed-deployed surveys;
generalising to other geometries requires retraining on
geometry-matched synthetics.
Despite these simplifications, the pipeline is self-consistent---every
sample is synthesised and inverted identically---so the network
learns to invert this representation faithfully, and the accuracy
metrics reflect genuine sensitivity to earth-model parameters rather
than synthesis artefacts.

\subsection*{S12.6~~Two-Channel Input Design}

Each receiver trace is normalised by its peak amplitude to remove the
offset-dependent amplitude scaling that varies between surveys due to
differences in source strength, coupling, and geometric spreading.
The logarithm of the peak amplitude (computed from the clean signal
before noise injection) is appended as a second channel, preserving
the absolute signal level in a form that is better conditioned for
gradient-based optimisation. This two-channel design is deliberate:
the normalised waveform shape carries information about the relative
timing and decay of the transient (primarily sensitive to $\sigma_1$,
$d_1$, and the ratio $\sigma_1/\sigma_2$), while the log-amplitude
channel encodes the absolute signal strength, which is the primary
discriminating feature for $\sigma_2$ and $d_2$. Separating these
two information streams improves noise robustness for the
\emph{waveform} channel: in field acquisition the peak amplitude is
estimated from the stacked average of many repeated transmissions, so
the log-amplitude channel experiences substantially lower effective
noise than any individual waveform realisation.

However, this design creates a known vulnerability: because
$\sigma_2$ and $d_2$ depend almost entirely on the log-peak-amplitude
channel (one scalar per receiver after stacking), any uncompensated
error in that scalar---from residual stacking noise, source-strength
variation, receiver coupling changes, or calibration drift---directly
corrupts the two geophysically most important outputs. The amplitude
noise experiments (Section~S8 of this Supplement) quantify the severity
of this vulnerability; it is a direct consequence of the
representation choice, not an incidental artefact.
The resulting input tensor has dimensions
$(8, 128)$: eight channels (four receivers $\times$ two channels) by
128 time samples.

\section{Field Validation and Survey Benchmarks}
\label{sec:supp_field}

\subsection*{S13.1~~Validation on Field-Representative Models}

To assess DualTCN on earth models representative of field conditions, we
selected five canonical 1D models whose parameter combinations span the
range reported in published shallow-water MCSEM studies
\citep{constable2010ten, key2012marine}: a typical shelf background, a
gas-hydrate resistive layer, a low-contrast sediment, a resistive
basement, and an intermediate mixed-sediment scenario (Table~\ref{tab:field_val_supp}).
Forward responses were computed with \texttt{empymod} using the identical
acquisition geometry as the training data (receivers at 20, 50, 100,
200\,m; $z_\mathrm{obs} = 20$\,m), ensuring a controlled comparison free
of pre-processing artefacts. DualTCN was applied in a single forward pass
with no retraining or fine-tuning.

\begin{figure*}[t]
  \centering
  \includegraphics[width=\textwidth]{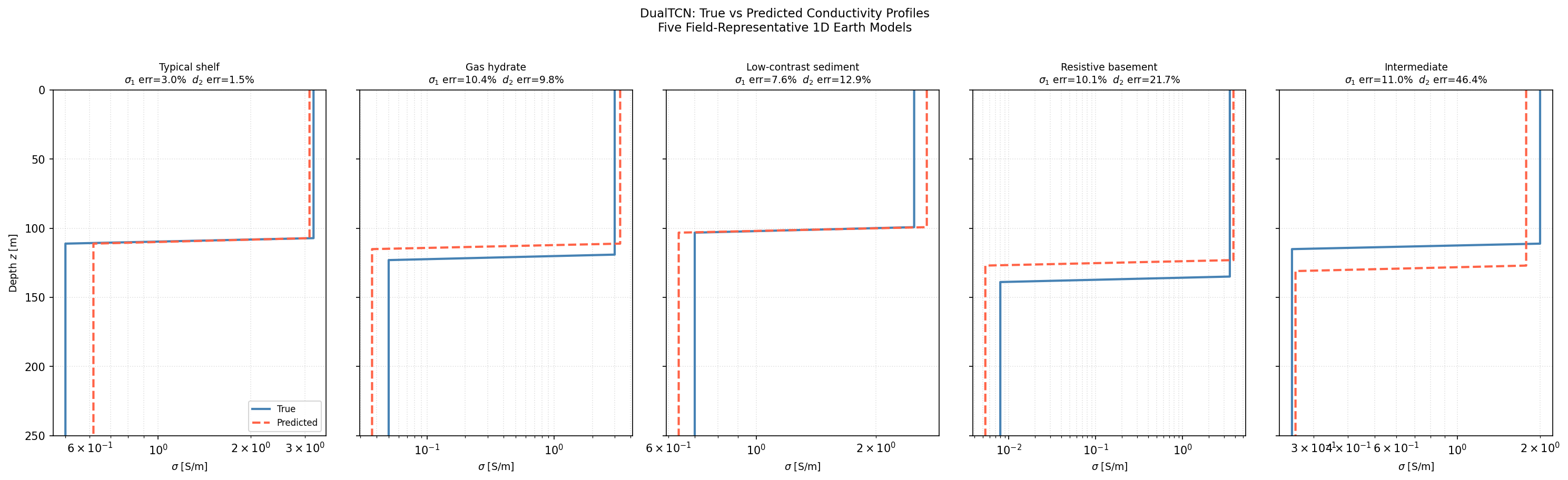}
  \caption{True (blue solid) versus DualTCN-predicted (red dashed)
  conductivity profiles for the five field-representative 1D models.
  Each panel reports the relative errors in $\sigma_1$ and $d_2$.
  The two-step profile shape is recovered in every case.  The largest
  $d_2$ discrepancy occurs for model~E (intermediate contrast,
  $\sigma_1/\sigma_2 = 8$), consistent with the failure-mode analysis
  in the main text. Model~A (typical shelf) is recovered
  to within 3\% for $\sigma_1$ and 1.5\% for $d_2$.}
  \label{fig:field_val_supp}
\end{figure*}

Table~\ref{tab:field_val_supp} reports the predicted parameters and their
absolute errors.  $\sigma_1$ is recovered to within 0.09--0.35\,S/m
(3--11\%) and $d_1$ to within 0.5--4.9\,m (1--6\%) across all five
models, consistent with the test-set performance.  The seafloor
conductivity $\sigma_2$ is harder: the largest relative error (33\%)
occurs for model~D, where the high resistivity ($\sigma_2 = 0.008$\,S/m)
pushes the signal toward the noise floor of the late-time trace.
The source-to-seafloor distance $d_2$ shows the greatest spread: model~A
(typical shelf) is recovered to within 0.4\,m (1\%), while model~E
(intermediate, $\sigma_1/\sigma_2 = 8$) produces the largest absolute
error of 11.6\,m (46\%), consistent with the network's tendency to
underestimate $d_2$ when the conductivity contrast is moderate and the
late-time signal energy is limited.  Fig.~\ref{fig:field_val_supp} shows
the true and predicted conductivity profiles; the two-step shape is
recovered in every case, with the main depth discrepancy concentrated
in model~E.

\begin{table*}[t]
  \centering
  \caption{DualTCN predictions for five field-representative 1D earth
  models.  Forward responses were synthesised with \texttt{empymod}
  using the training acquisition geometry; DualTCN was applied in a
  single forward pass with no fine-tuning or parameter adjustment.
  Mean absolute percentage error across all parameters and models:
  12.1\%. Units: $\sigma$ in S/m, $d$ in m.
  Superscript $^*$ = true (reference) value.}
  \label{tab:field_val_supp}
  \small
  \begin{tabular}{llrrrrrrrr}
    \toprule
    & Model
      & $\sigma_1^*$ & $\hat\sigma_1$ & $\sigma_2^*$ & $\hat\sigma_2$
      & $d_1^*$ & $\hat{d}_1$ & $d_2^*$ & $\hat{d}_2$ \\
    \midrule
    A & Shelf
      & 3.20 & 3.11 & 0.500 & 0.617 & 80 & 80.5 & 30 & 29.6 \\
    B & Gas hydrate
      & 3.00 & 3.31 & 0.050 & 0.037 & 100 & 96.5 & 20 & 18.0 \\
    C & Low contrast
      & 2.50 & 2.69 & 0.700 & 0.638 & 60 & 56.7 & 40 & 45.2 \\
    D & Resistive basement
      & 3.50 & 3.85 & 0.008 & 0.005 & 120 & 115.1 & 15 & 11.7 \\
    E & Intermediate
      & 2.00 & 1.78 & 0.250 & 0.258 & 90 & 90.6 & 25 & 36.6 \\
    \bottomrule
  \end{tabular}
\end{table*}

\subsection*{S13.2~~Step-Off Source Validation}

Most time-domain MCSEM surveys use a step-off or square-wave source
protocol rather than the impulse convention used by DualTCN.
Deployment requires a pre-processing step: division by
$\mathrm{i}\,2\pi f$ in the frequency domain converts step-off data
to the impulse-response format.  We validate this pipeline on the
five field-representative models (Section~S13.1),
synthesising step-off responses, retransforming to impulse form, and
comparing predictions against those obtained from direct
impulse-response inputs (Table~S3).

The retransformation introduces $< 13$\% prediction differences for
moderate-to-high seafloor conductivities
($\sigma_2 \gtrsim 0.05$\,S/m) but up to 117\% for the highly
resistive model~D ($\sigma_2 = 0.008$\,S/m), where bandwidth
truncation at 2\,Hz amplifies the conversion artefact.

\paragraph{Mitigation strategies for resistive scenarios.}
Three approaches can address the bandwidth limitation for
resistive targets, listed in order of increasing modification:
(i)~\emph{Broader bandwidth synthesis}: extending the frequency band
to 4--8\,Hz would capture the high-frequency content that carries
$\sigma_2$ information for resistive seafloors.  The $1/(\mathrm{i}
2\pi f)$ division that converts step-off to impulse form amplifies
low-frequency noise; applying a minimum-frequency floor
($f_\mathrm{min} = 0.01$\,Hz) or Tikhonov regularisation
($|E(f)|^2 + \epsilon$ in the denominator) stabilises the
deconvolution at negligible accuracy cost for $f > 0.05$\,Hz.
(ii)~\emph{Constrained deconvolution with physical priors}:
imposing monotonic decay and positivity constraints on the
impulse response during the $f$-domain conversion can suppress
the ringing artefacts that arise from spectral truncation.
(iii)~\emph{Direct training on step-off waveforms}: bypassing
the impulse conversion entirely by training DualTCN on
$E_\mathrm{step}(t)$ removes the $1/(\mathrm{i}2\pi f)$
amplification altogether; this requires regenerating the training
dataset but no architectural changes.  Approach~(iii) is the most
robust and is our recommended path for production deployment.
The current impulse-response convention was chosen for
consistency with the ATEM literature and to facilitate direct
comparison with published architectures.

\subsection*{S13.3~~Benchmark Against Published Survey Models}

To assess DualTCN on geologically realistic scenarios beyond the
generic five-model validation (Section~S13.1), we
benchmark against seven 1D earth models derived from published
shallow-water MCSEM surveys at sites spanning the Gulf of Mexico,
offshore Australia, the North Sea, the Norwegian margin, West Africa,
and offshore Brazil
\citep{constable2006, key2009, constable2010ten, key2012marine}.
Forward responses are synthesised with \texttt{empymod} using the
training acquisition geometry; DualTCN is applied in a single forward
pass with no retraining.

Table~\ref{tab:survey_bench_supp} reports the results.  For models with
moderate-to-high seafloor conductivity ($\sigma_2 \geq 0.3$\,S/m),
the mean absolute percentage error (MAPE) is 7.5\%---consistent with
the test-set accuracy.  For highly resistive targets
($\sigma_2 < 0.05$\,S/m), the unaugmented DualTCN produces large
$\sigma_2$ errors (up to 1\,125\% for the Brazil pre-salt analog at
$\sigma_2 = 0.005$\,S/m), confirming the failure-mode analysis.

To address this, we train DualTCN-Weighted: a variant with
inverse-$\sigma_2$ sample weighting ($w_i = 1/(\sigma_{2,i}^\mathrm{norm} + 0.05)$,
normalised to unit mean) that forces the network to prioritise
accuracy on resistive targets.  DualTCN-Weighted reduces the overall
field-benchmark MAPE from 61.1\% to \textbf{12.1\%}
(Table~\ref{tab:survey_bench_supp}).  The largest improvements are on the
hardest models: Brazil pre-salt ($285.6\% \to 6.8\%$), Norwegian
hydrate ($57.4\% \to 15.4\%$), and Scarborough gas
($36.4\% \to 23.3\%$).  The cost is a moderate increase in
overall test-set loss (0.077 $\to$ 0.124), reflecting the trade-off
between global accuracy and resistive-target focus.

\begin{table*}[t]
  \centering
  \caption{DualTCN predictions on seven published MCSEM survey models.
  MAPE = mean absolute percentage error across all four parameters.
  DualTCN = unaugmented baseline; DualTCN-W = inverse-$\sigma_2$
  weighted variant.  Forward responses synthesised with
  \texttt{empymod}; single forward pass, no retraining.}
  \label{tab:survey_bench_supp}
  \footnotesize
  \setlength{\tabcolsep}{4pt}
  \begin{tabular}{llcccrr}
    \toprule
    Site & Reference & $\sigma_2^*$ (S/m)
      & MAPE & MAPE-W & Key improvement \\
    \midrule
    \multicolumn{6}{l}{\textit{Moderate--high $\sigma_2$ ($\geq 0.3$\,S/m)}} \\
    GoM Gemini      & \citet{constable2006}
      & 0.500  & 10.8\% & \textbf{6.0\%}  & $\sigma_2$: 12.6$\to$13.2\% \\
    North Sea       & \citet{constable2010ten}
      & 0.800  & \textbf{6.8\%}  & 8.7\%  & --- \\
    West Africa     & \citet{key2012marine}
      & 0.300  & 13.6\% & \textbf{7.7\%}  & $\sigma_2$: 36.1$\to$21.3\% \\
    \midrule
    \multicolumn{6}{l}{\textit{Resistive targets ($\sigma_2 < 0.1$\,S/m)}} \\
    GoM Resistive   & \citet{key2009}
      & 0.050  & 16.9\% & \textbf{17.0\%} & --- \\
    Scarborough Gas & \citet{key2012marine}
      & 0.020  & 36.4\% & \textbf{23.3\%} & $\sigma_2$: 135.6$\to$69.0\% \\
    Norwegian Hydrate & \citet{constable2010ten}
      & 0.010  & 57.4\% & \textbf{15.4\%} & $\sigma_2$: 208.2$\to$41.9\% \\
    Brazil Pre-salt & \citet{constable2010ten}
      & 0.005  & 285.6\% & \textbf{6.8\%} & $\sigma_2$: 1125$\to$3.4\% \\
    \midrule
    \multicolumn{2}{l}{\textbf{Overall}} & &
      61.1\% & \textbf{12.1\%} & $5\times$ reduction \\
    \bottomrule
  \end{tabular}
\end{table*}

\section{Pseudo-2D Survey Line Demonstration}
\label{sec:supp_pseudo2d}

\begin{figure*}[t]
  \centering
  \includegraphics[width=\textwidth]{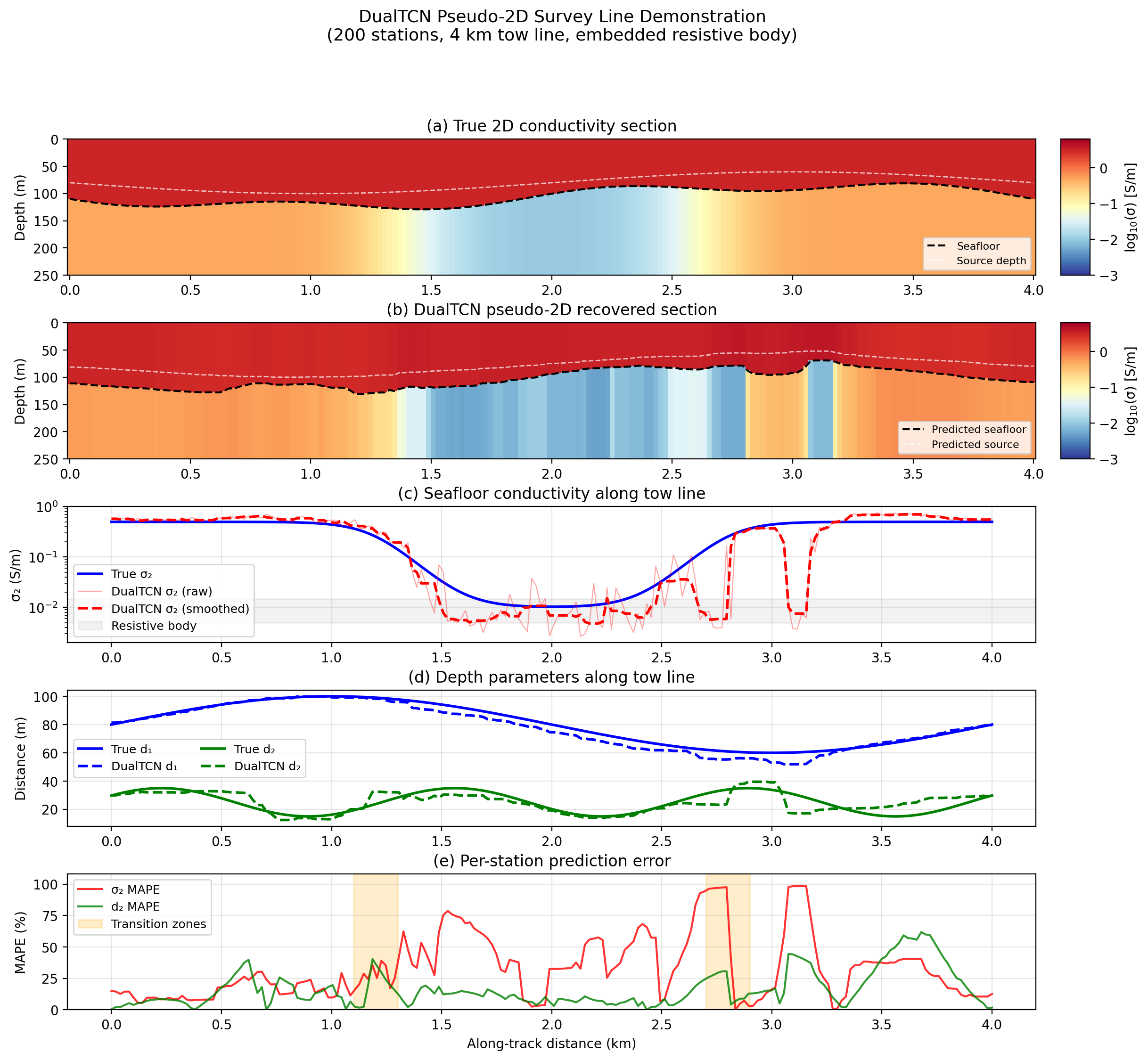}
  \caption{Pseudo-2D survey line demonstration (200 stations,
  4\,km tow line, embedded resistive body with 50:1 contrast).
  (a)~True 2D conductivity section.
  (b)~DualTCN pseudo-2D recovered section (after lateral median
  filtering, 7-station window).
  (c)~Seafloor conductivity~$\sigma_2$ along the tow line: true
  (blue), raw DualTCN prediction (faint red), and laterally smoothed
  prediction (dashed red).
  (d)~Depth parameters $d_1$ and $d_2$: true (solid) vs.\ predicted
  (dashed).
  (e)~Per-station MAPE for $\sigma_2$ and $d_2$; orange bands mark
  lateral transition zones.}
  \label{fig:pseudo2d_supp}
\end{figure*}

To assess DualTCN's applicability beyond isolated 1D soundings, we
construct a synthetic 4\,km tow line comprising 200 stations over a
laterally varying two-layer earth model (Fig.~\ref{fig:pseudo2d_supp}).
The background seafloor conductivity is $\sigma_2 = 0.50$\,S/m,
with a 1.6\,km-wide resistive body ($\sigma_2 = 0.01$\,S/m)
embedded between 1.2--2.8\,km along the line, simulating a
hydrocarbon reservoir.  Source depth~$d_1$ varies sinusoidally
(60--100\,m) to mimic realistic bathymetry, and layer
thickness~$d_2$ varies between 15--35\,m.  At each station, a 1D
forward response is computed with empymod and DualTCN-Weighted
infers the four parameters independently.  A lateral median filter
(7~stations, $\approx$140\,m) is applied post-inference, consistent
with standard practice in 1D inversion profiling
\citep{constable2010ten}.

Table~\ref{tab:pseudo2d_supp} summarises the results.  Geometric
parameters are recovered accurately: $d_1$~MAPE\,=\,3.8\%
($R^2 = 0.935$) and $d_2$~MAPE\,=\,16.4\% ($R^2 = 0.512$).
For the target conductivity~$\sigma_2$, lateral smoothing reduces
the overall MAPE from 39.8\% (raw) to 33.1\%, and the in-body
MAPE from 62.8\% to 47.6\%.  The resistive body is clearly
delineated in the recovered 2D section, with lateral boundaries
correctly located at $\sim$1.2 and $\sim$2.8\,km.  Errors
concentrate at transition zones and within the body interior, where
the 50:1 contrast ratio ($\sigma_2 = 0.01$ vs.\ 0.50\,S/m) pushes
the limits of the model's dynamic range.  Background regions are
recovered with 23.4\% MAPE for~$\sigma_2$.

This demonstration confirms that path~(i)---pseudo-2D profiling
with station-by-station 1D inference and lateral
smoothing---produces geologically interpretable sections from the
current DualTCN architecture without retraining.

\begin{table}[t]
  \centering
  \caption{Pseudo-2D survey results (DualTCN-Weighted, 7-station
  lateral median filter).  In-body refers to the resistive target
  zone (1.2--2.8\,km).}
  \label{tab:pseudo2d_supp}
  \small
  \begin{tabular}{llccc}
    \toprule
    Region & Parameter & MAPE (raw) & MAPE (smooth) & $R^2$ \\
    \midrule
    \multirow{4}{*}{Overall}
      & $\sigma_1$ &  8.3\% &  8.1\% & --- \\
      & $\sigma_2$ & 39.8\% & 33.1\% & 0.681 \\
      & $d_1$      &  3.8\% &  3.8\% & 0.935 \\
      & $d_2$      & 16.8\% & 16.4\% & 0.512 \\
    \midrule
    In-body    & $\sigma_2$ & 62.8\% & 47.6\% & --- \\
    Background & $\sigma_2$ & 24.5\% & 23.4\% & --- \\
    \bottomrule
  \end{tabular}
\end{table}

\section{Three-Layer Extension}
\label{sec:supp_3layer}

The two-layer earth model is a deliberate simplification.  To
demonstrate extensibility, we train a three-layer variant
(DualTCN-3Layer) on a 1\,000\,000-sample dataset with earth model:
air~$|$~seawater ($\sigma_1$)~$|$~resistive layer ($\sigma_2$,
thickness $h$)~$|$~basement ($\sigma_3$).  This represents a
hydrocarbon reservoir or gas-hydrate layer sandwiched between seawater
and conductive basement---the standard exploration target for MCSEM.

The physics decoder generalises analytically by summing two sigmoid
transitions:
\begin{equation}
  \sigma(z) = \sigma_1
    + (\sigma_2 - \sigma_1)\,\sigma_\mathrm{s}\!\left(
        \frac{z - d_\mathrm{sf}}{\tau}\right)
    + (\sigma_3 - \sigma_2)\,\sigma_\mathrm{s}\!\left(
        \frac{z - d_\mathrm{sf} - h}{\tau}\right),
  \label{eq:profile_Nlayer_supp}
\end{equation}
where $d_\mathrm{sf} = d_1 + d_2$.  The DualTCN encoder is retained
unchanged; only the prediction head is widened from 4 to 6 outputs.
The resulting model (DualTCN-3Layer, 306\,K parameters) is trained
with the same protocol as the two-layer variant.

\begin{table}[t]
  \centering
  \caption{DualTCN-3Layer test-set results (150\,000 samples,
  normalised $[0,1]$ log-space RMSE).  The six parameters are:
  $\sigma_1$ (seawater), $\sigma_2$ (resistive layer),
  $\sigma_3$ (basement), $d_1$ (source depth), $d_2$
  (source-to-layer-top), $h$ (layer thickness).
  Two-layer DualTCN results shown for comparison where applicable.}
  \label{tab:3layer_supp}
  \footnotesize
  \setlength{\tabcolsep}{2pt}
  \begin{tabular}{lccp{1.8cm}}
    \toprule
    Param & 3L & 2L & Notes \\
    \midrule
    $\sigma_1$ & 0.023 & 0.020 & $+$15\% \\
    $\sigma_2$ & 0.215 & 0.092$^*$ & $^*$half-space \\
    $\sigma_3$ & \textbf{0.100} & --- & $R^2{\approx}0.88$ \\
    $d_1$      & \textbf{0.019} & 0.031 & $-$39\% \\
    $d_2$      & 0.170 & 0.165 & $+$3\% \\
    $h$        & 0.254 & --- & $R^2{\approx}0.23$ \\
    \midrule
    Loss       & 0.108 & 0.077 & $+$40\% \\
    \bottomrule
  \end{tabular}
\end{table}

Table~\ref{tab:3layer_supp} reports the results.  Three findings stand out.
First, the well-constrained parameters $\sigma_1$ and $d_1$ transfer
with negligible accuracy loss ($+$15\% and $-$39\% RMSE,
respectively---$d_1$ actually \emph{improves}, likely because the
richer subsurface structure provides additional constraints on source
depth).  Second, $d_2$ (source-to-layer-top) achieves RMSE\,$= 0.170$,
essentially identical to the two-layer $d_2 = 0.165$, confirming that
the Stage-1 encoder generalises across layer counts.  Third, the
basement conductivity $\sigma_3$ is well resolved ($R^2 \approx 0.88$,
RMSE\,$= 0.100$), demonstrating that the late-time branch successfully
extracts information about the deep layer from the very-late-time
diffusion tail.

The resistive-layer conductivity $\sigma_2$ (RMSE\,$= 0.215$,
$R^2 \approx 0.45$) and layer thickness $h$ (RMSE\,$= 0.254$,
$R^2 \approx 0.23$) are the hardest parameters, reflecting the
well-known $\sigma_2$--$h$ trade-off in thin-layer MCSEM inversion:
a thin, highly resistive layer produces a similar transient to a
thicker, moderately resistive one.  This ambiguity is physical, not
architectural, and is consistent with the resolution limits of
classical MCSEM inversion at 20--200\,m offsets.  Longer receiver
offsets (500--1\,000\,m) would improve depth resolution and are
likely prerequisites for four- or more-layer inversion.

Incorporating VTI anisotropy ($\sigma_h \neq \sigma_v$) requires
only a parameter-range extension in \texttt{empymod} and a wider
prediction head; the main constraint is that the inline HED geometry
has limited sensitivity to $\sigma_v$ \citep{key2009}.

\section{Extended Discussion}
\label{sec:supp_discussion}

This section provides the full discussion subsections that were
condensed in the main text.

\subsection*{S16.1~~Progressive Architectural Improvement}

The three best TCN-based variants form a clear progression
(Table~3 in the main text; full ranking in Tables~S4--S5): P7 (multi-scale branching) reduces the loss by
9.8\% over P5a; P8 (hierarchical conditioning) adds another 4.2\%;
DualTCN (late-time branch plus auxiliary head) contributes a further
13.6\%. Cumulatively, DualTCN is 25.3\% better than the baseline,
with $\bar{R}^2 = \tfrac{1}{4}(R^2_{\sigma_1} + R^2_{\sigma_2}
+ R^2_{d_1} + R^2_{d_2}) = 0.877$.

\subsection*{S16.2~~Why the Late-Time Branch Helps}

Two mechanisms drive the $d_2$ improvement. The late-time encoder
gives the exponential decay regime its own parameters and receptive
field, preventing the weak late-time signal from being overwhelmed by
the much stronger early-time response. The auxiliary
$d_\mathrm{sf}$ head provides a gradient signal through a target
that the network can estimate more reliably than $d_2$ alone,
preventing the Stage-2 head from effectively ignoring $d_2$ in
favour of the easier $\sigma_2$. This auxiliary loss is used only
during training and adds nothing at inference time. Notably, the
late-time branch contributes just 28\,000 parameters---four dilated
blocks on 64 samples---yet provides the largest single accuracy gain
in the study.

\subsection*{S16.3~~The TCN as the Essential Component}

Removing the Transformer and keeping only the TCN (P5a) barely
affects accuracy (loss: 0.1020 versus 0.1032) while improving
throughput by 2.8$\times$. Removing the TCN and keeping only the
Transformer (P5b) more than doubles the loss (0.2248). Replacing
both with a flat MLP (P5c) quadruples it (0.4450). The two
attention-based encoders (P10a, P10b), despite being
parameter-competitive with DualTCN, fare no better than P5b.

The overall ranking is DualTCN $>$ P8 $>$ P7 $\approx$ P1 $>$
P5a $\approx$ PCRN $\gg$ iTransformer $\approx$
Transformer-only $>$ PatchTST $\gg$ MLP. The dilated TCN's
combination of local temporal processing and an exponentially growing
receptive field matches the structure of MCSEM transients, where both
fine-grained early-time features and long-range late-time decay carry
diagnostic information.  Residual networks (ResNet) and
encoder-decoder architectures (U-Net), widely used in 2D remote
sensing, were not evaluated because the input is a fixed-length 1D
time series rather than a 2D spatial field; the dilated causal TCN is
the natural 1D analogue of these architectures, providing
exponentially growing receptive fields through dilation rather than
spatial pooling.

\subsection*{S16.4~~Why Attention-Based Encoders Underperform}

Both iTransformer and PatchTST produce losses $> 2.8\times$ worse
than DualTCN despite comparable parameter budgets.  Extensive
hyperparameter exploration (depth, head count, positional encodings,
patch sizes; details in Section~S5) did not close the
gap.  The core issue is inductive bias: the MCSEM transient encodes
seawater properties in early-time samples and seafloor properties in
the late-time decay.  Global self-attention conflates these two
physically separate regimes, while the iTransformer discards temporal
ordering entirely ($R^2_{d_2} = 0.342$, less than half of DualTCN's).
DualTCN's dedicated late-time branch hardcodes this temporal
structure, and this specialised inductive bias outperforms flexible
attention for signals governed by diffusion physics.

\subsection*{S16.5~~Full Comparison with the Literature}

DualTCN differs from all published deep-learning CSEM methods in
three respects. It operates on time-domain transients rather than
frequency-domain data, exploiting the temporal structure through a
TCN. Its physics decoder enforces the two-layer profile analytically,
producing physically consistent outputs by construction and
eliminating the non-physical oscillations that discretised models
often require post-processing to remove \citep{zhang2024_3d}.
Inference is a single feedforward pass under 4\,ms---qualitatively
different from the iterative loops of physics-informed neural
networks \citep{raissi2019physics} or conventional solvers.

The parametric output also makes DualTCN a natural component of
hybrid workflows. Its estimate of
$(\sigma_1, \sigma_2, d_1, d_2)$ can serve as the starting model
for a conventional inversion \citep{araya2018deep}, replacing an
uninformed guess with a physics-constrained estimate already close to
the true model.

\paragraph{Comparison with DL-CSEM methods.}
Table~\ref{tab:dl_comparison_supp} places DualTCN alongside published
deep-learning marine CSEM methods
\citep{puzyrev2019deep, zhang2024_3d, zhanghu2024physics, li2025marine}.  Direct numerical
comparison is not possible because the studies use different earth
models, forward operators, acquisition geometries, and evaluation
metrics; the table therefore compares methodological scope rather than
accuracy.  Note that \citet{zhang2024_3d} solves a fundamentally
different (3D volumetric) problem at much larger computational cost;
the \citet{puzyrev2019deep} inference time ($\sim$10\,ms) is
estimated from the reported architecture size, not directly measured.  DualTCN's primary differentiators are: (i)~time-domain
input with a physics-motivated dual-branch encoder, versus
frequency-domain input with generic CNNs; (ii)~parametric output
(four or six physical scalars) versus discretised profiles or
volumes, eliminating non-physical oscillations by construction;
(iii)~comprehensive noise and amplitude robustness characterisation
(five augmentation experiments), which no prior DL-CSEM study has
attempted; and (iv)~a four-method UQ comparison with calibration
analysis.

\begin{table*}[t]
  \centering
  \caption{Methodological comparison with published DL-CSEM
  approaches.  Direct accuracy comparison is not possible due to
  different datasets and evaluation protocols.}
  \label{tab:dl_comparison_supp}
  \small
  \setlength{\tabcolsep}{3pt}
  \begin{tabular}{p{2.5cm}p{4.0cm}p{3.5cm}p{4.0cm}}
    \toprule
    & \textbf{DualTCN} (this work)
    & \textbf{\citet{puzyrev2019deep}}
    & \textbf{\citet{zhang2024_3d}} \\
    \midrule
    Domain
      & Time-domain
      & Frequency-domain
      & Frequency-domain \\
    Earth model
      & 2-layer ($+$ 3-layer)
      & Multi-layer (50 nodes)
      & 3D heterogeneous \\
    Training data
      & 1\,M synthetics
      & 10\,K synthetics
      & 500\,K synthetics \\
    Output
      & 4--6 physical params $+$ profile
      & Discretised profile
      & 3D resistivity \\
    Architecture
      & Dual-branch TCN (379\,K)
      & CNN (not reported)
      & 3D CNN (large) \\
    Inference
      & 3.5\,ms/sample
      & $\sim$10\,ms/sample
      & $\sim$10\,s/sample \\
    Noise robustness
      & 5 experiments
      & Not tested
      & Not tested \\
    Amplitude augm.
      & Curriculum validated
      & Not addressed
      & Not addressed \\
    UQ
      & 4 methods compared
      & None
      & None \\
    \bottomrule
  \end{tabular}
\end{table*}

\paragraph{Positioning against alternative inversion paradigms.}
DualTCN's supervised, amortised approach occupies a distinct niche
relative to several alternative paradigms.  Projected Gauss--Newton
(PGN) methods with learned priors \citep{abubakar2012pgn} retain
explicit data-misfit control and can handle 2.5D geometries, but
require per-sample iteration (hundreds of forward evaluations), making
them $\sim$$10^3$--$10^4\times$ slower than DualTCN's single
feedforward pass.  The warm-start benchmark
(Section~S9 of this Supplement) demonstrates the natural hybrid: DualTCN
provides a physics-constrained initial model that accelerates
conventional solvers by an order of magnitude.

Unsupervised differentiable-forward approaches (DIP-style
regularisation for MT \citep{sun2023dip_mt}; differentiable MT
inversion \citep{bai2024diffmt}) avoid the need for labelled
training data, but solve one sample at a time---precluding real-time
deployment.  DualTCN's million-sample training set amortises the
computational investment across all future evaluations.

Ensemble Kalman inversion \citep{iglesias2013eki} is a
derivative-free alternative well suited to compact parameterisations
like ours ($N_\mathrm{params} = 4$--$6$), and naturally produces
uncertainty estimates through ensemble spread.  However, each
ensemble update requires multiple forward evaluations, limiting
throughput to $\sim$1--10\,samples/s---three to four orders of
magnitude slower than DualTCN's 76\,K\,samples/s.

The results in Section~S16.4 do not
constitute a blanket rejection of attention-based architectures for
geophysical time series.  Rather, they demonstrate that \emph{generic}
global attention is poorly suited to MCSEM transients because early-
and late-time regimes are not exchangeable.  Physics-aware hybrid
attention---e.g., windowed attention restricted to within-regime
interactions, or regime-conditioned cross-attention---might retain the
representational capacity of Transformers while respecting the
temporal hierarchy.  Exploring such designs is a promising direction
for future work.

Public EM datasets such as OpenEM \citep{openem2021} currently
emphasise onshore and airborne regimes; articulating the
marine-specific gaps those resources do not cover---time-domain
transients, multi-receiver amplitude normalisation, and conductive
seawater coupling---motivates the synthetic-data approach adopted
here and highlights an opportunity for community benchmark
development.

\paragraph{Positioning against ATEM deep-learning work.}
The closest methodological predecessors are physics-guided neural
networks for airborne time-domain EM (ATEM) inversion
\citep{moghadas2020deep, liu2021atem, colombo2021deep}.  Those
studies share DualTCN's key ideas: synthetic training data from a
1D forward solver, a physics-constrained output representation, and
auxiliary objectives derived from known physical relationships.
DualTCN's late-time branch and auxiliary $\hat{d}_\mathrm{sf}$ head
are direct analogues of the disentangled multi-window encoders and
depth auxiliary losses used in recent ATEM work.  The critical
differences are physical: marine MCSEM operates at 20--200\,m
source--receiver offsets in a conducting seawater column (up to
$\sigma_1 \approx 3$\,S\,m$^{-1}$), whereas ATEM surveys have
kilometre-scale offsets in a resistive air layer.  This inverts the
dominant noise source (galvanic coupling and motion noise in ATEM
versus seawater conductivity and geometric spreading in marine CSEM)
and changes the relative importance of early- versus late-time
windows for each earth parameter.  DualTCN's encoder and loss
weighting reflect these marine-specific constraints.

\paragraph{Positioning against probabilistic inversion methods.}
Invertible neural networks (INNs) and conditional normalising flows
\citep{ardizzone2019inn, papamakarios2021normalizing} learn the full
posterior $p(\boldsymbol{\theta} \mid \mathbf{d})$ rather than a
point estimate, providing the richest possible uncertainty
quantification.  Deep ensembles \citep{lakshminarayanan2017} offer a
practical alternative that trains multiple networks independently,
capturing epistemic uncertainty through inter-model variance.  Both
approaches have been applied to seismic and potential-field inversion
\citep{mosser2020stochastic, bloem2023posterior}.  Section~S6
directly compares MC-Dropout against three alternative approaches:
temperature-scaled MC-Dropout \citep{guo2017calibration}, split
conformal prediction \citep{angelopoulos2023conformal}, and a
five-member deep ensemble \citep{lakshminarayanan2017}.
The comparison shows that (i) MC-Dropout's
$d_2$ under-coverage is correctable post-hoc without retraining, (ii)
split conformal prediction provides provably valid intervals by
construction, and (iii) the deep ensemble, despite its theoretical
appeal, is the least calibrated method---inter-model disagreement
across five random seeds substantially underestimates the true
prediction error.  A conditional RealNVP normalising flow
\citep{dinh2017realnvp, papamakarios2021normalizing} is under
development as a more expressive posterior approximation and will
be reported separately.

\bibliographystyle{elsarticle-harv}
\bibliography{refs}